\PassOptionsToPackage{numbers}{natbib}

\documentclass{article}

    \usepackage[preprint]{neurips_data_2023}

\usepackage[utf8]{inputenc} 
\usepackage[T1]{fontenc}    
\usepackage{hyperref}       
\usepackage{url}            
\usepackage{booktabs}       
\usepackage{amsfonts}       
\usepackage{nicefrac}       
\usepackage{microtype}      
\usepackage{xcolor}         

\usepackage{graphicx}
\usepackage{comment}
\usepackage{amsmath}
\usepackage{color}
\usepackage{subcaption}
\usepackage{booktabs}
\usepackage{multirow}
\usepackage{array}
\usepackage{wrapfig}
\usepackage{enumitem}

\definecolor{corner}{RGB}{33,102,172}
\definecolor{sidewalk}{RGB}{239,138,98}
\definecolor{crossing}{RGB}{178,24,43}
\definecolor{classtext}{RGB}{253,219,199}

\definecolor{Red}{RGB}{225, 25, 5}
\definecolor{Green}{RGB}{51, 160, 44}
\definecolor{Blue}{RGB}{0, 100 ,200}
\definecolor{Yellow}{RGB}{255, 255 ,0}

\definecolor{BrightRed}{RGB}{225, 0, 0}
\definecolor{LightYellow}{RGB}{254, 254 ,139}

\title{PathwayBench: Assessing Routability of Pedestrian Pathway Networks Inferred from Multi-City Imagery}

\author{%
  Yuxiang Zhang, Bill Howe, Sachin Mehta, Nicholas-J Bolten, Anat Caspi \\
  University of Washington\\
  \texttt{\{yz325,billhowe,sacmehta,bolten,caspian\}@uw.edu} \\
}

\begin{document}

\maketitle

\begin{abstract}
Applications to support pedestrian mobility in urban areas require an accurate, complete, and routable graph representation of the built environment. Globally available information, including aerial imagery provides a scalable, low-cost source for constructing these path networks, but the associated learning problem is challenging: Relative to road network pathways, pedestrian network pathways are narrower, more frequently disconnected, often visually and materially variable in smaller areas (as opposed to roads' consistency in a region or state), and their boundaries are broken up by driveway incursions, alleyways, marked or unmarked crossings through roadways.  Existing algorithms to extract pedestrian pathway network graphs are inconsistently evaluated and tend to ignore routability, making it difficult to assess utility for mobility applications: Even if all path segments are available, discontinuities could dramatically and arbitrarily shift the overall path taken by a pedestrian. In this paper, we describe a first standard benchmark for the pedestrian pathway network graph extraction problem, comprising the largest available dataset equipped with manually vetted ground truth annotations (covering $3,000 km^2$ land area in regions from 8 cities), and a family of evaluation metrics centering routability and downstream utility.  By partitioning the data into polygons at the scale of individual intersections, we can compute local routability as an efficient proxy for global routability.  We consider multiple measures of polygon-level routability, including connectivity, degree centrality, and betweenness centrality, and compare predicted measures with ground truth to construct evaluation metrics. Using these metrics, we show that this benchmark can surface strengths and weaknesses of existing methods that are hidden by simple edge-counting metrics over single-region datasets used in prior work, representing a challenging, high-impact problem in computer vision and machine learning.
\end{abstract}

\section{Introduction}
\label{sec:intro}


\begin{figure}[t!]
\centering
\includegraphics[width=0.75\columnwidth]{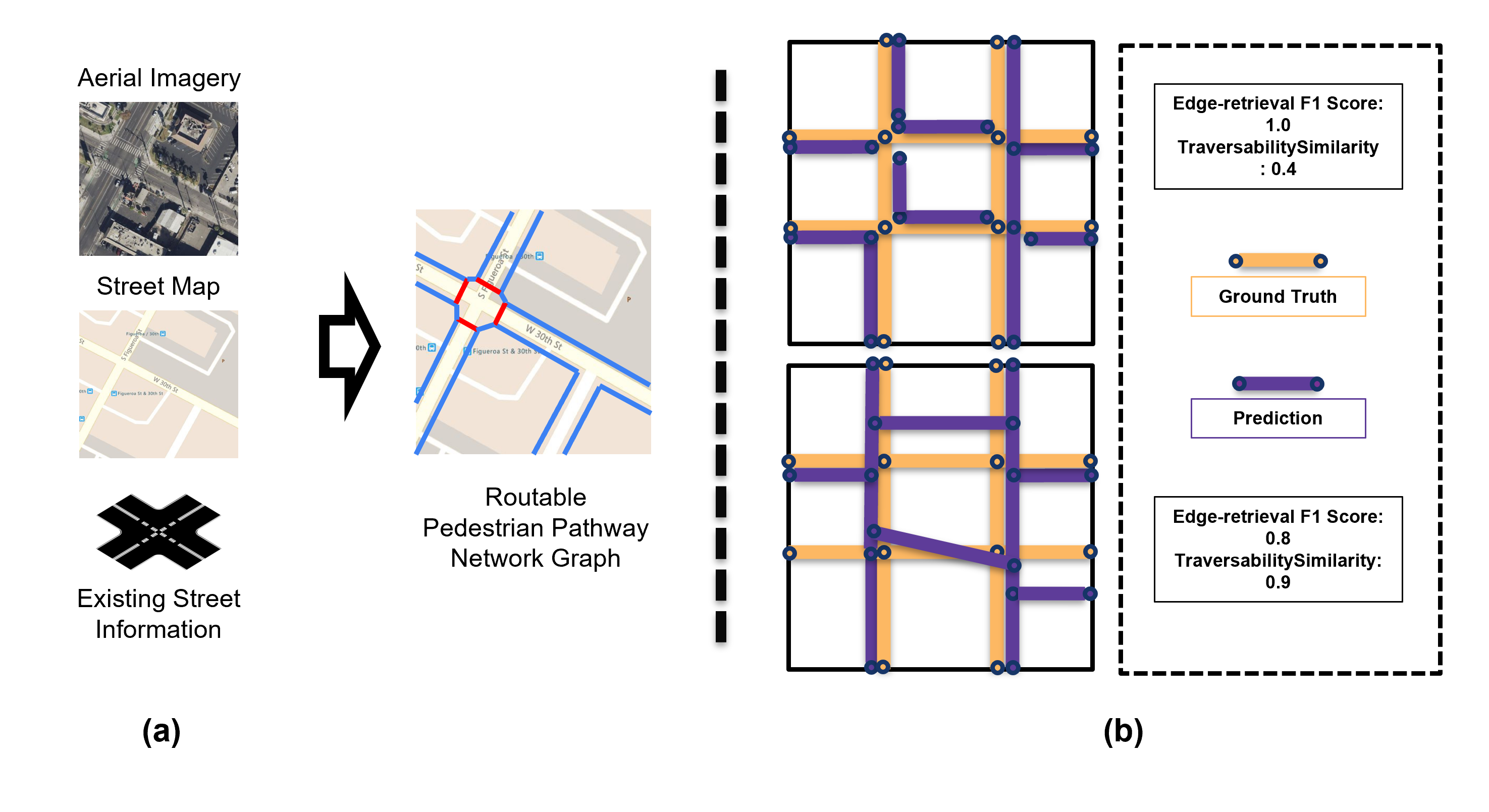}
\caption{  (a) PathwayBench evaluates solutions to the \textbf{pedestrian pathway inference problem}: Given a dataset of co-registered aerial images, street map images, and road networks (Section \ref{sec:dataset}), produce a routable pedestrian network graph. (b) A new metric of local \textbf{{traversability}} (Section \ref{sec:tiletraversability}) provides an efficient proxy for global graph routability that centers the traveler experience.}
\label{fig:flow_chart}
\vspace{-6mm}
\end{figure}

A routable pedestrian path network detailing the location and connectivity of sidewalks, crossings, and curbs is essential for both city planning tasks and wayfinding applications~\cite{bolten2017pedestrian, bolten2022towards}.  While on-the-ground mapping provides high-quality results, it is prohibitively expensive for use in multiple large metropolitan areas.  Automatic extraction of accurate, routable paths from remote sensing inputs is therefore a foundational spatial analysis problem in mobility data acquisition ~\cite{hosseini2023mapping, zhang2023ape}. 

While roadways have been extensively mapped~\cite{mattyus2017deeproadmapper, mi2021hdmapgen}, pedestrian paths are increasingly a priority for city planning while representing a fundamentally more challenging learning problem, for several reasons. First, the components of a routable pedestrian path are highly variable: They may consist of sidewalks, marked and unmarked crossings in roadways, curb ramps, and other transitions~\cite{bolten2021towards}.  Second, the paths themselves may built from different materials with different visual properties in aerial imagery (concrete, paving stones, or asphalt). Third, the boundaries of pedestrian paths are less consistent, having variable widths with interruptions by, for example, driveway incursions or alleyways.  Fourth, ground truth for pedestrian paths is less reliable: Even with training, human mappers introduce geometry-based errors that can complicate validation~\cite{bolten2022towards}.  

These challenges have motivated new approaches specialized to pedestrian paths rather than adaptations of methods for extracting road networks~\cite{karimi2013pedestrian, li2018semi,hosseini2023mapping, zhang2023ape}. However, the complexity of the problem has led to ad hoc evaluation procedures, with each method using different urban areas (with different properties), different assumptions of the availability of features, different sources of ground truth, and evaluation using different metrics (Table \ref{tab:related}).



In this work, we present PathwayBench 
, comprising 1) a multi-city dataset with externally validated features and output and 2) an evaluation procedure emphasizing graph connectivity and therefore utility in downstream applications. 
The dataset includes regions from three cities (Seattle, Portland, and Washington D.C.), representing two relatively similar, smaller cities in the Pacific Northwest and one larger city on the East Coast. For each city region, the dataset includes five co-registered, externally validated features: aerial imagery, road networks (with corresponding rasterized street map tiles that include additional annotations), and human-validated pedestrian pathways (with corresponding rasterized pathways for evaluating segmentation methods). The evaluation procedure comprises a set of typical, standardized metrics for both the rasterized annotations and the pathway graph, as well as a novel traversability metric that captures global routability in a local, efficient computation. We show that this metric can expose quality issues in extracted graphs that conventional edge-counting metrics do not, and therefore provide a better basis for assessing utility for downstream applications.
Figure \ref{fig:flow_chart} illustrates the problem setting (a) and the motivation for traversaibility (b).

We make the following contributions:
\begin{itemize}[leftmargin=*]
\item We provide a human-validated ground truth pathway graph for three city regions representing diverse built environments, along with co-registered rasterizations of these graphs to support the evaluation of segmentation-based vision methods. 
\item We present a set of evaluation criteria emphasizing graph connectivity, including a new metric that uses the traversability of small regions as an efficient proxy for computing global path preservation in a complex urban pedestrian environment.
\item We present an evaluation of state-of-the-art methods for the pedestrian pathway extraction problem, demonstrating that the diverse dataset exposes unique challenges and that the proposed metrics expose quality issues that correlate with downstream utility. 
\end{itemize}


The paper is organized as follows. Section \ref{sec:related} discusses related work. Section \ref{sec:dataset} introduces the PathwayBench dataset. Section \ref{sec:tiletraversability} describes our metrics for partition-level graph routability analysis. Experiment results are shown in Section \ref{sec:exp}. Discussion and conclusions are presented in Section \ref{sec:dis}.

\section{Related Work}
\label{sec:related}
We discuss related work in three domains:studies on street/road network mapping, methods to map pedestrian environments,  and available remote sensing datasets. The combined work highlights the need for a standard dataset and benchmark that targets the pedestrian pathway graphs.   

\begin{minipage}[t!]{0.99\textwidth}
    \centering
    \captionsetup{width=0.99\textwidth}
    \captionof{table}{Related methods, assumed inputs, outputs, and evaluation method.  }
    \label{tab:related}
    \resizebox{0.99\textwidth}{!}{%
        \begin{tabular}{l|c|c|c|c}
        \toprule
        {\textbf{Method}}
        &  \textbf{Inputs to Method} & \textbf{Method Outputs} & \textbf{Evaluation Method} & \textbf{Evaluation Data Area}\\
        \midrule
        \textbf{MD-ResUNet \cite{wu2019road}} & VHR satellite imagery & Road Segmentation & mIoU, precision, F1 & Seat, America\\ 
        \textbf{Li et al. \cite{li2018semi}} & Parcel-level data, roadway centerline data & Pathway network & Count of sidewalk and crosswalk features & Atlanta\\
        \textbf{Tile2Net \cite{hosseini2023mapping}} & Orthorectified aerial imagery & Segmentation, pathway network & mIoU, edge-retrieval recall & Cambridge, Boston, Manhattan \\
        \textbf{Pedestrainfer \cite{zhang2023ape}} & Road network & Pathway network & Edge-retrieval precison, recall, F1 & Los Angeles, Bellevue, Quito \\
        \textbf{Prophet \cite{zhang2023ape}} & Road network, rasterized street map, aerial imagery & Segmentation, pathway network & mIoU, edge-retrieval precison, recall, F1 & Los Angeles, Bellevue, Quito\\
        \bottomrule
        \end{tabular}
    }
\end{minipage}

\textbf{Street and Road Network Map Generation}
Studies have investigated using aerial imagery along with auxiliary data for street mapping. Wu et al.  \cite{wu2019road} used OpenStreetMap (OSM) centerlines as labeled data and extracted roads from very high-resolution (VHR) satellite images. Sun et al. \cite{sun2019leveraging} added crowd-sourced global positioning system (GPS) data to satellite images to extract roads with CNN-based semantic segmentation. Zhou et al. \cite{zhou2021funet} fused remote sensing images and GPS for road detection and extraction. Additional recent learning-based studies included Lu et al.  \cite{lu2021gamsnet} proposing a multi-scale residual neural architecture for road detection, Pan et al. \cite{pan2021generic} proposing a fully convolutional neural network using VHR remote sensing, Mattyus et al. \cite{mattyus2017deeproadmapper} estimated road topology from aerial images, Mi et al. \cite{mi2021hdmapgen} generated road lane graphs from LiDAR data with a hierarchical graph generation model. Importantly, work in this domain solely focuses on automobile road detection and extraction, and does not address the generalization or extension of the proposed methods to the pedestrian environment.  Methods for mapping the environments that serve pedestrians' travel have not been widely studied. 

\textbf{Mapping the Pedestrian Environment}
Few studies have focused on mapping pedestrian environments compared to automobile roads. Karimi et al. \cite{karimi2013pedestrian} explored pedestrian map generation approaches in a small-scale area, demonstrating preliminary mapping results that heavily depended on the availability and quality of input data. Recent advancements in remote sensing have led to more imagery-based approaches, such as Ahmetovic et al. \cite{ahmetovic2015zebra} detection of zebra crossings using satellite imagery and validation with street-level images. Likewise, Ghilardi et al. \cite{ghilardi2016crosswalk} classified and located crosswalks using an SVM classifier over data extracted from road maps, and Ning et al. \cite{ning2022sidewalk} extracted sidewalks from aerial images with a neural network and restored occluded segments from street view images. These studies improved pedestrian environment mapping, but they do not generate a comprehensive, connected, and routable pedestrian pathway network graph needed for city planning and navigation. Other studies use on-the-ground data, such as Zhang et al. \cite{zhang2021collecting} automated collection of street-view images with auxiliary data to map sidewalk connectivity and sidewalk infrastructure. Hou et al. \cite{hou2020network} extract sidewalk paths using LiDAR data and point cloud segmentation. These methods often require physical systems to cover a large area to generate a pedestrian pathway network. Other studies that map pedestrian pathway networks at scale are often based solely on existing street (road) data. For example, Li et al. \cite{li2018semi}'s semi-automated method generated a sidewalk network using parcel-level data and roadway centerline, but it required human editing for quality control. 
The Pedestrianfer system \cite{zhang2023ape} uses existing street network graph information to automate the pathway network generation, but it only provides an optimistic hypothesis from incomplete information and does not capture the true connectivity.
Recently, the Tile2Net system \cite{hosseini2023mapping} uses satellite imagery to segment sidewalks, crosswalks, and footpaths in cities, then simplifies the segmented polygons and extracts sidewalk centerlines. This work highlights the challenges of feature detection from only satellite imagery because of the vegetation obstructions over the footpaths, and the difficulties in correctly representing the path connectivity and routability when fitting centerlines to discrete polygons. The Prophet system \cite{zhang2023ape} uses the existing road networks, and segmentation from aerial imagery and rasterized street maps to generate pathway network graphs. 
We show the performance of Tile2Net and Prophet using our benchmark in Section \ref{sec:eval_graph}.

\textbf{Remote Sensing Datasets}
There are several remote sensing datasets that are used for mapping. The TorontoCity dataset \cite{wang2016torontocity} contains aerial satellite images for road curb extraction and road centerline estimation. PRRS \cite{aksoy2008performance} presents a dataset for building extraction and Digital Surface Model (DSM) estimation using satellite data. In addition, the DeepGlobe dataset \cite{DeepGlobe18} and the ISPRS dataset \cite{ISPRS} contain imagery and annotations for tasks including road extraction, building detection, and land cover classification. These datasets enable researchers to study different tasks involving the use of aerial imagery data, but they do not provide pedestrian pathway graph annotations or a standard way to evaluate the pathway graph. Overall, no current dataset provides ground truth pedestrian pathway graphs with relevant features used in their extraction.

\section{The PathwayBench Dataset}
\label{sec:dataset}

To address the lack of large-scale datasets that target the pedestrian environment, we collect annotations of connected, annotated pedestrian path networks represented as graphs. We also provide rasterizations of the graphs into pixel-labeled images for assessing segmentation-based algorithms. In addition to these outputs, we provide co-registered aerial imagery, annotated road networks represented as graphs, and rasterized street maps as input features. These inputs and outputs collectively form the PathwayBench dataset.

\subsection{Coverage}
\label{sec:coverage}
The PathwayBench dataset, and our evaluation, center on select regions in three US cities: Seattle, Portland, and Washington DC.  The PathwayBench dataset also includes data from regions in additional cities: Bellevue, WA; Quito, Ecuador; Sao Paulo, Brazil; Santiago, Chile; and Valparaiso, Chile. We focus on US cities in this paper to ensure fair comparisons across multiple methods.


\subsection{Data Collection and Annotations}
\label{sec:collection}

The graph annotations for pedestrian pathways are represented in the GeoJSON format. The data is usually mapped and maintained by city agencies or crowdsourcing by local mappers through platforms like OpenStreetMap (OSM) \cite{haklay2008openstreetmap}. In each area included in the PathwayBench dataset, mappers from the OpenSidewalks Project \cite{bolten2017pedestrian, bolten2022towards} first delineated pathways according to the standardized OpenSidewalks data schema \cite{opensidewalksSchema}. Subsequently, each mapped feature underwent a validation process conducted by independent mappers to ensure accuracy and consistency. Each dataset sample (Figure \ref{fig:data_sample}) includes (1) the aerial imagery, (2) the road network, (3) the rasterized street map tiles that include additional annotations, (4) the human-validated pedestrian pathway graph, and (5) the rasterized pathways for supporting segmentation methods. The aerial imagery and rasterized street map tiles are acquired from Bing Maps along every major road for each area described in Section \ref{sec:coverage}. Each set of samples is precisely aligned and co-registered to the same bounding box geographically.


\subsection{Annotations to Evaluate Segmentation-based Methods}
\label{sec:classes}
\begin{figure}[ht!]
    \centering
    \resizebox{0.85\textwidth}{!}{
        \begin{tabular}{cccccc}
            \rotatebox{90}{\parbox{3cm}{\centering\textbf{Aerial imagery}}} & 
            \includegraphics[width=0.2\textwidth]{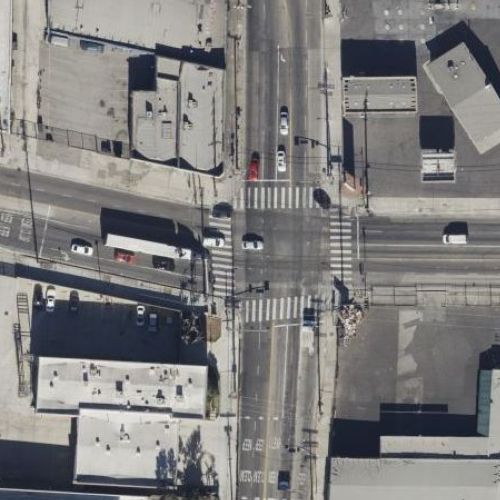} & 
            \includegraphics[width=0.2\textwidth]{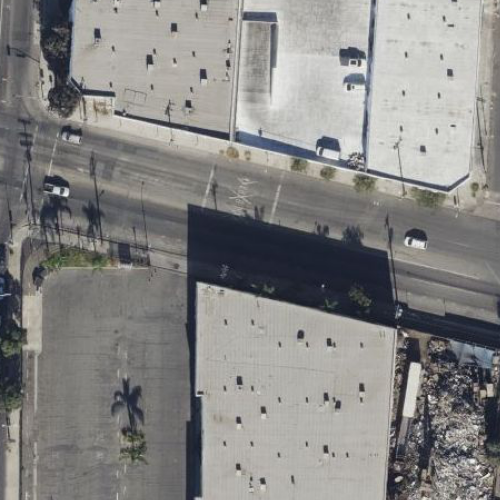} & 
            \includegraphics[width=0.2\textwidth]{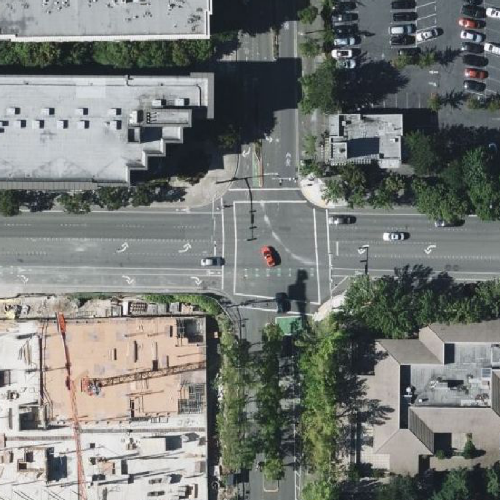} & 
            \includegraphics[width=0.2\textwidth]{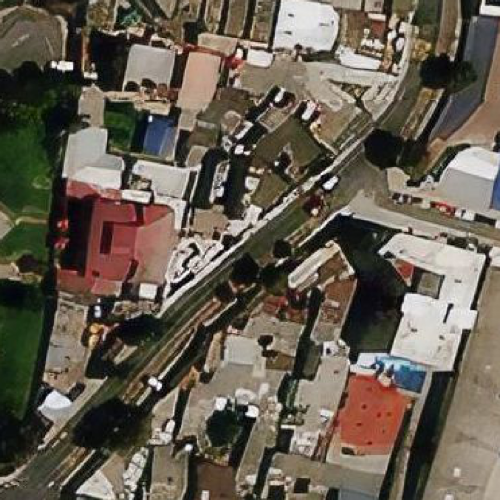} &
            \includegraphics[width=0.2\textwidth]{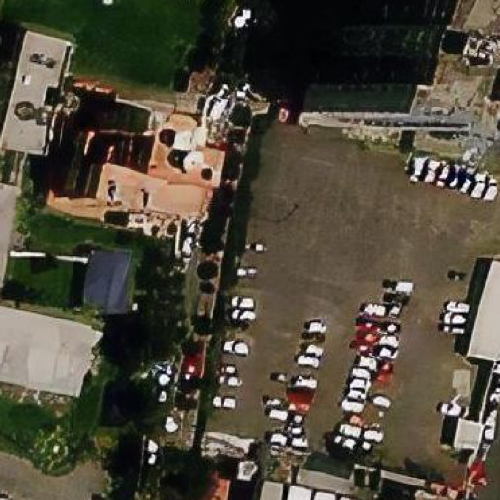} 
            \\
            \\
            \rotatebox{90}{\parbox{3cm}{\centering\textbf{Road network}}} & 
            \includegraphics[width=0.2\textwidth]{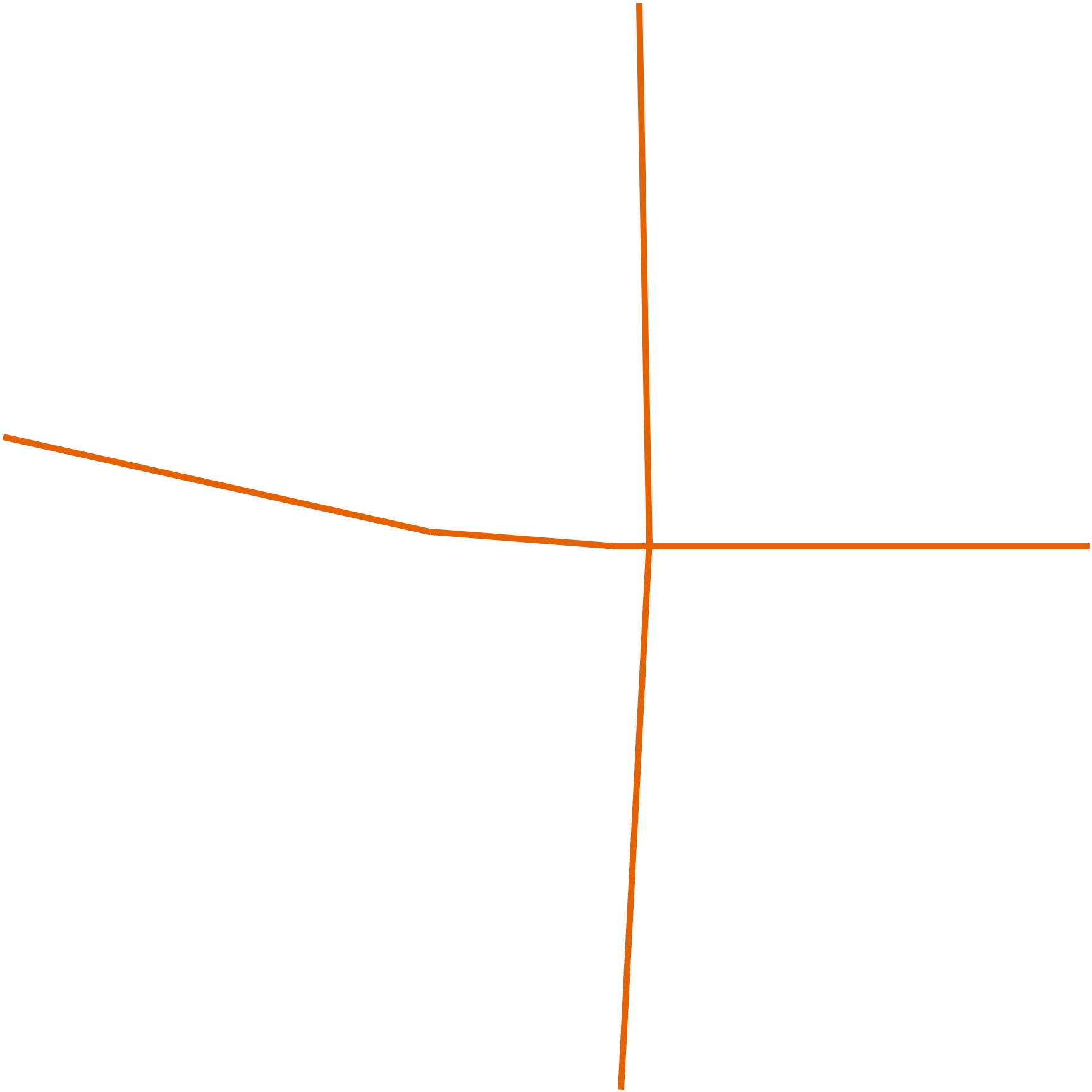} & 
            \includegraphics[width=0.2\textwidth]{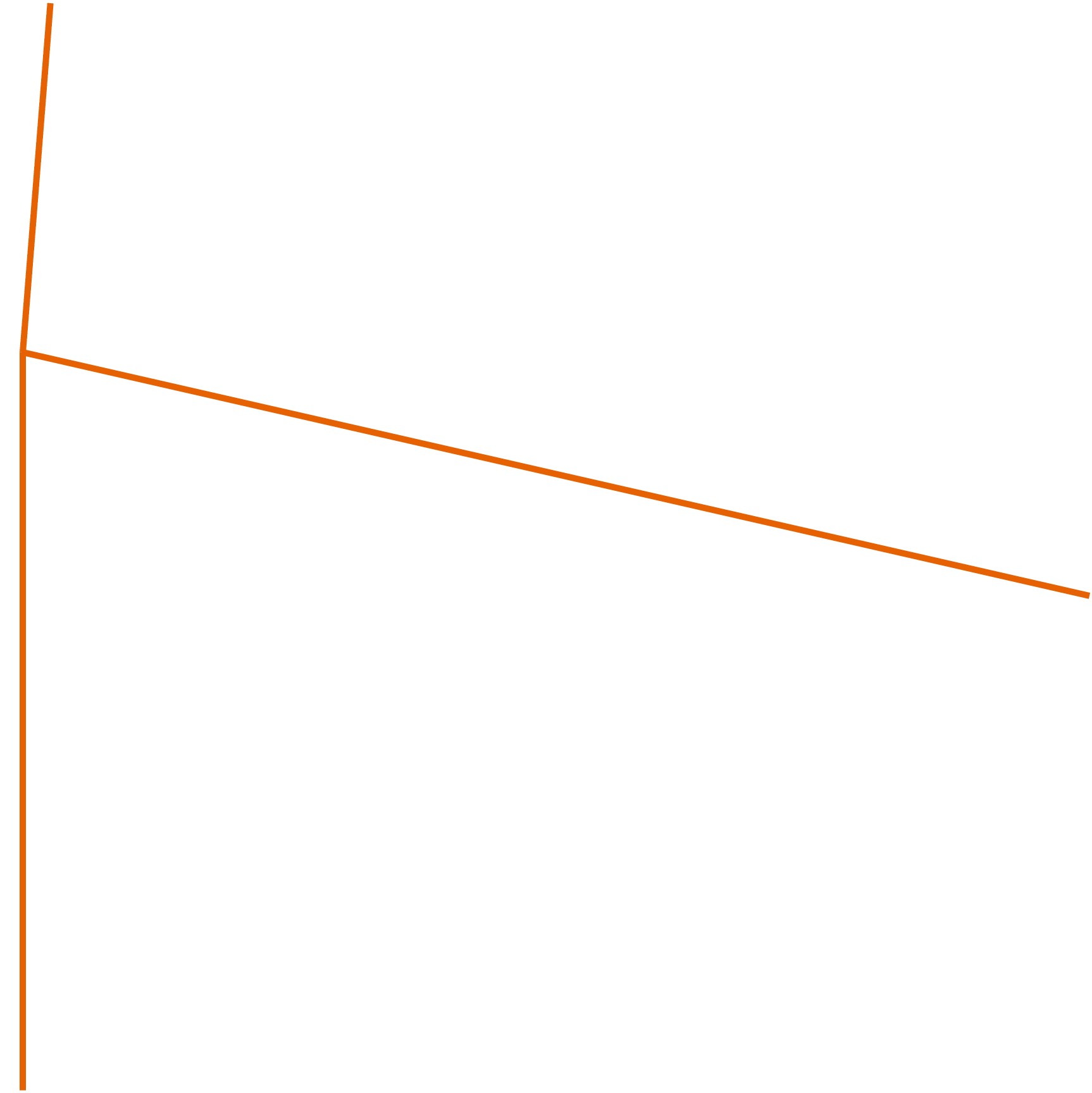} & 
            \includegraphics[width=0.2\textwidth]{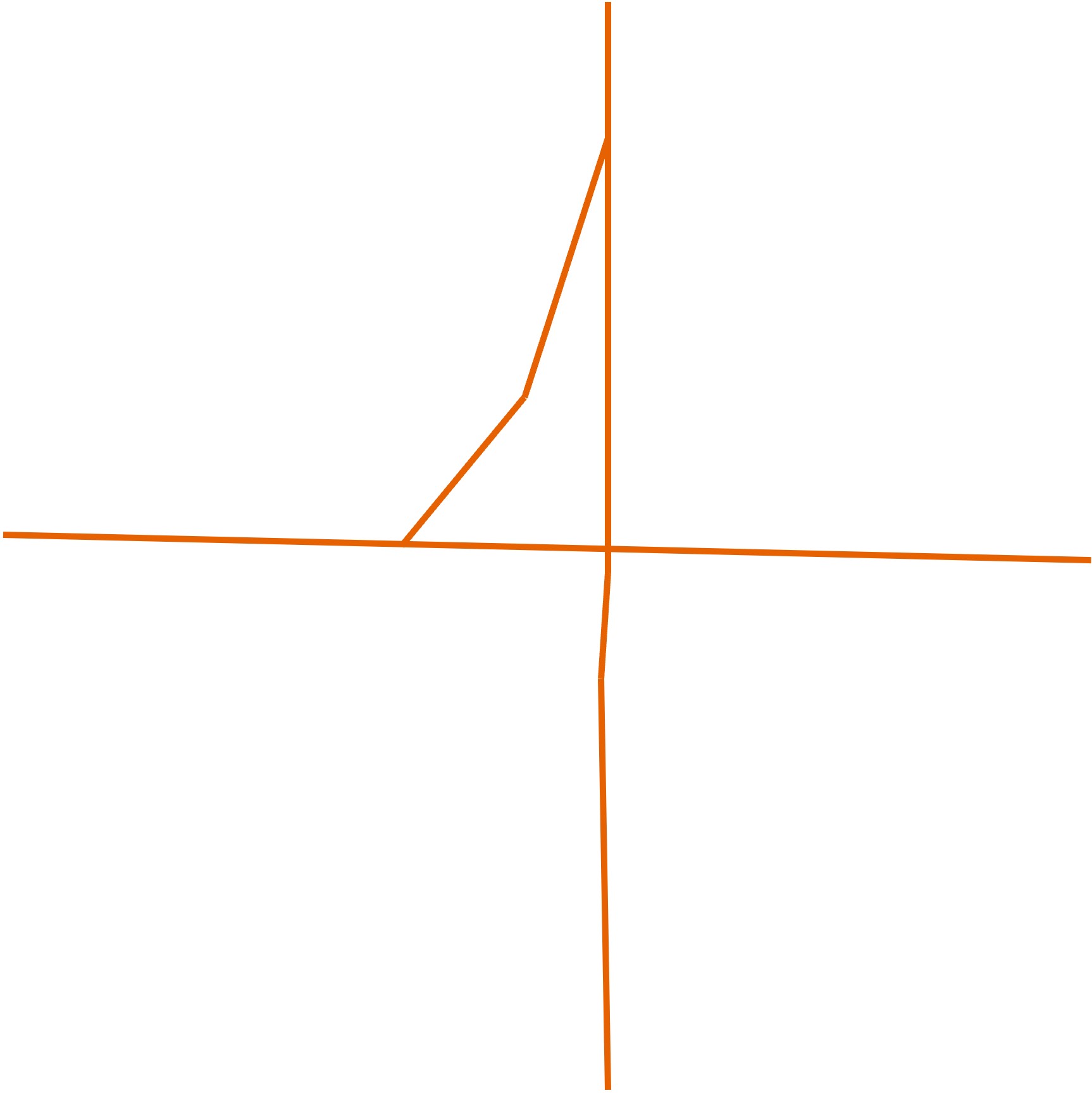} & 
            \includegraphics[width=0.2\textwidth]{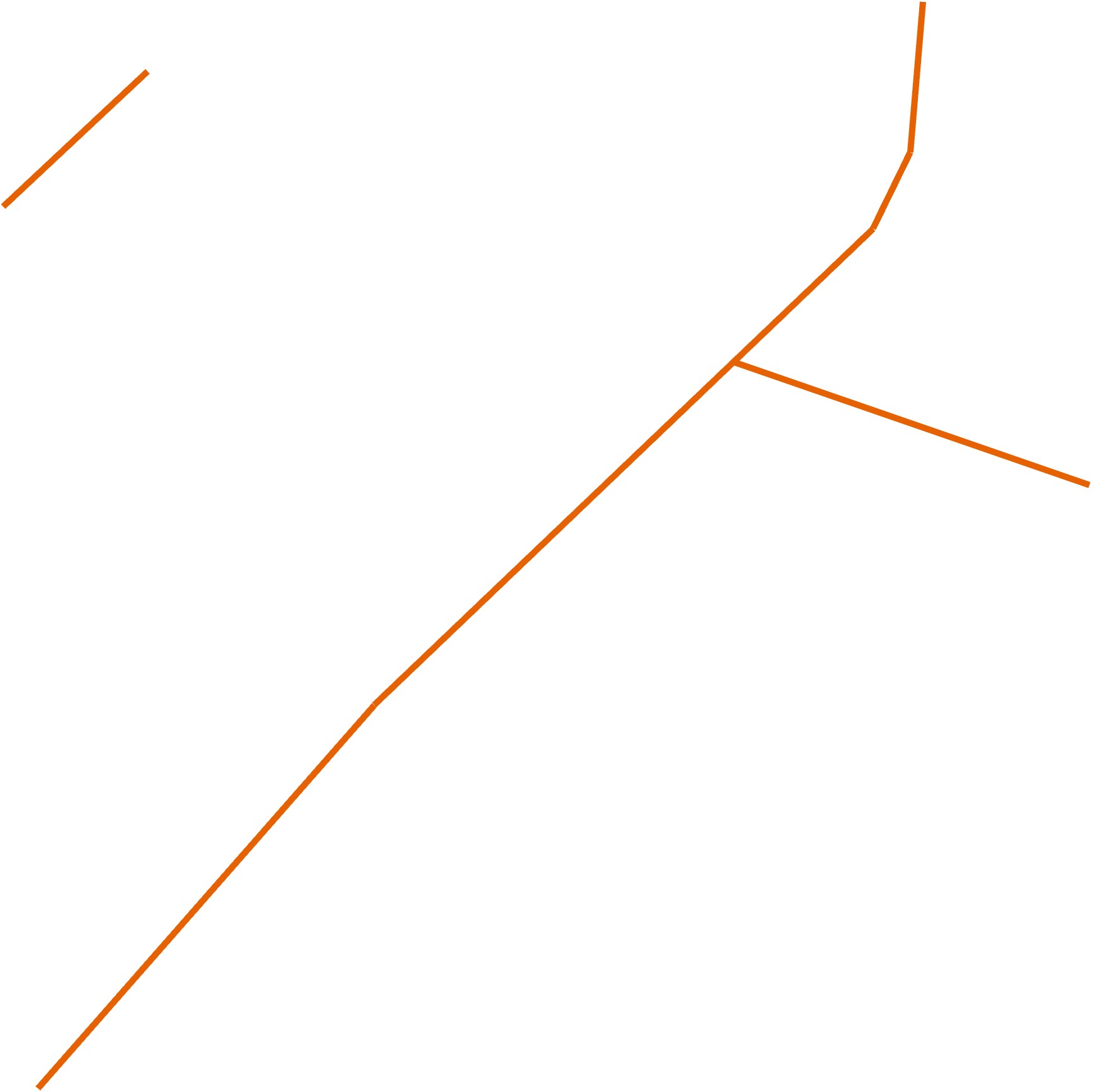} &
            \includegraphics[width=0.2\textwidth]{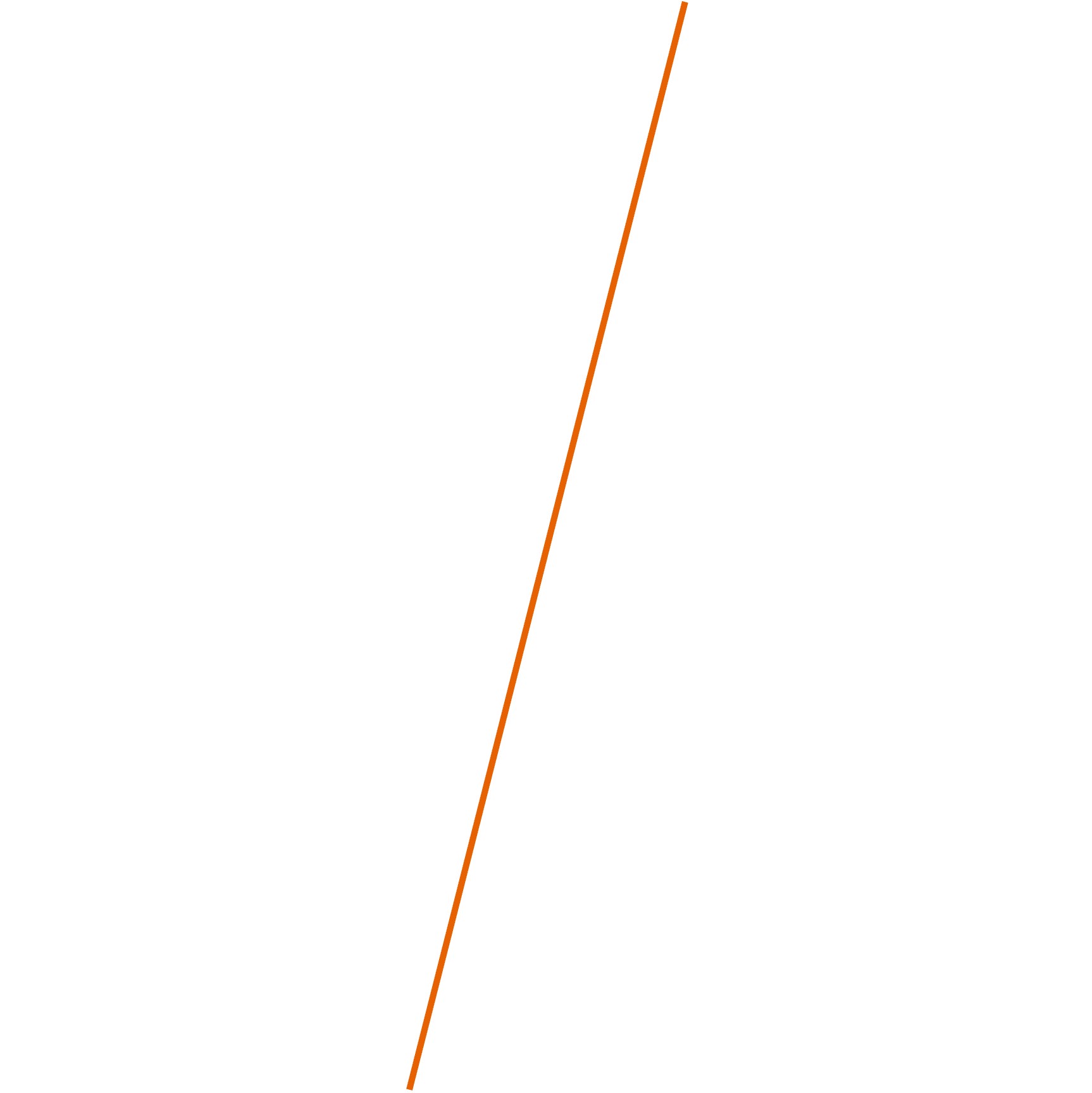}
            \\
            \\
            \rotatebox{90}{\parbox{3cm}{\centering\textbf{Rasterized\\street map}}} & 
            \includegraphics[width=0.2\textwidth]{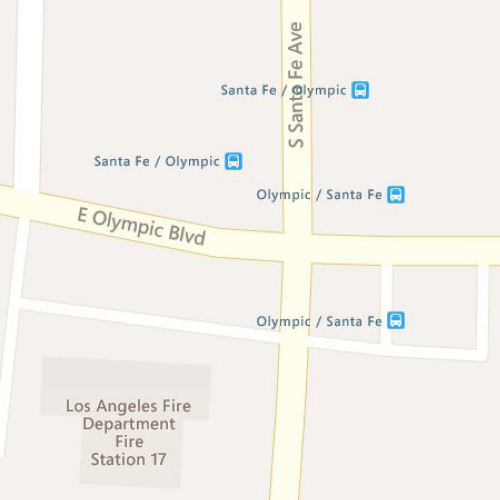} & 
            \includegraphics[width=0.2\textwidth]{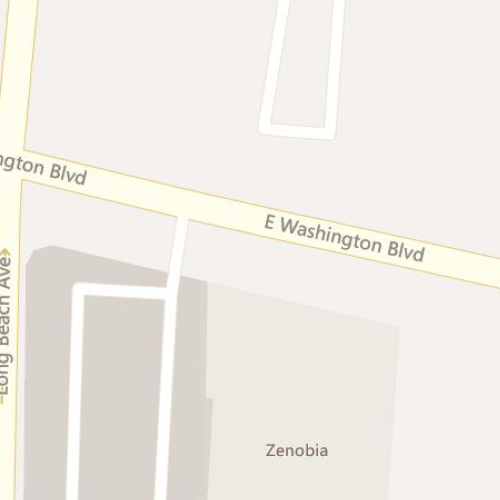} & 
            \includegraphics[width=0.2\textwidth]{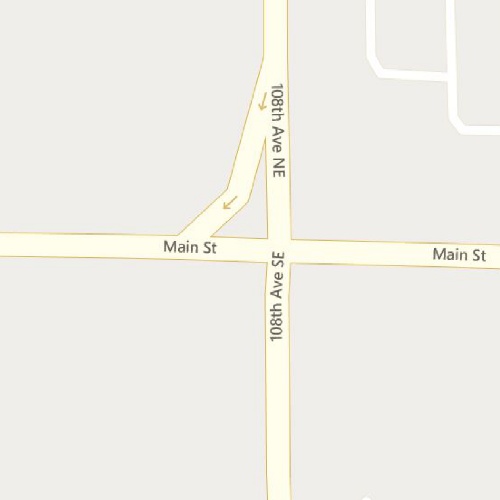} & 
            \includegraphics[width=0.2\textwidth]{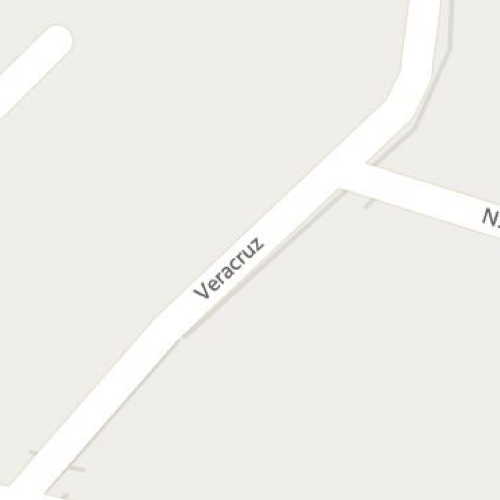} &
            \includegraphics[width=0.2\textwidth]{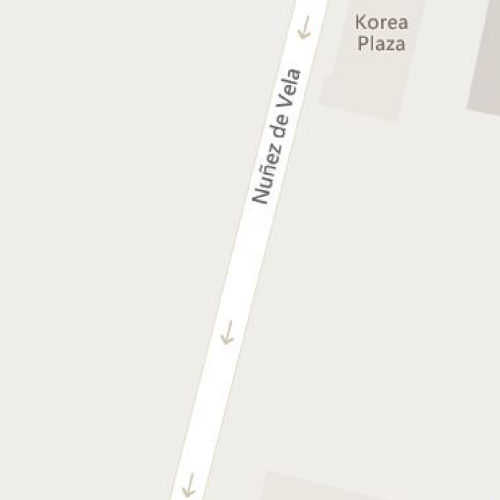}
            \\
            \\
            \rotatebox{90}{\parbox{3cm}{\centering\textbf{Rasterized\\pathway graph}}} & 
            \includegraphics[width=0.2\textwidth]{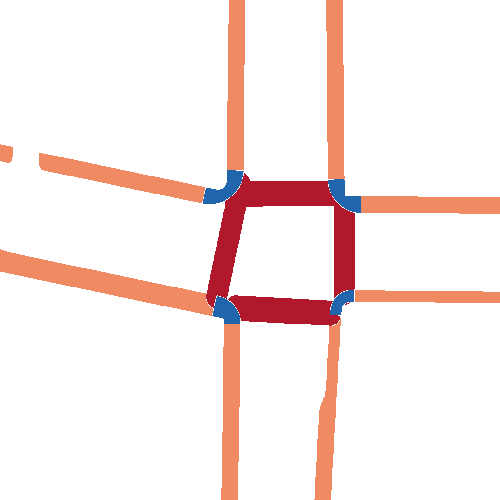} & 
            \includegraphics[width=0.2\textwidth]{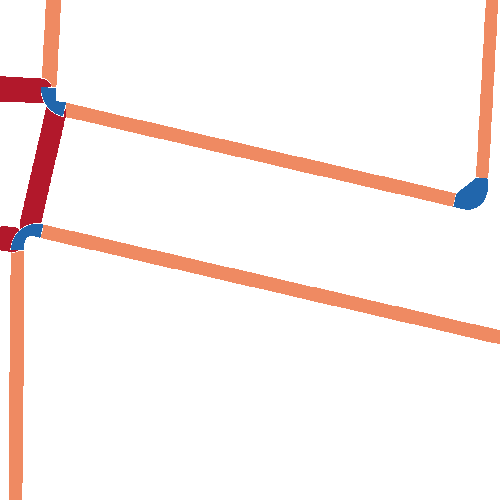} & 
            \includegraphics[width=0.2\textwidth]{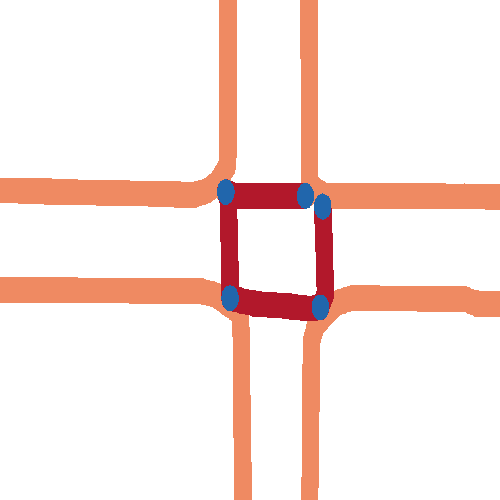} & 
            \includegraphics[width=0.2\textwidth]{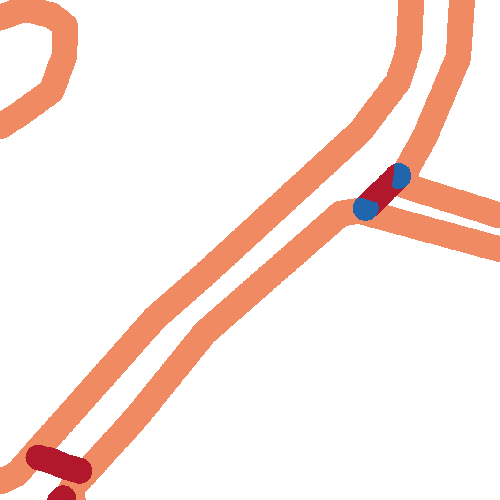} &
            \includegraphics[width=0.2\textwidth]{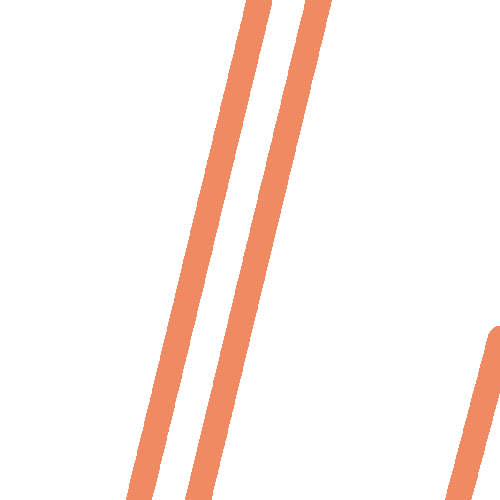} 
            \\
            \rotatebox{90}{\parbox{3cm}{\centering\textbf{Pedestrian\\pathway graph}}}
            & 
            \includegraphics[width=0.2\textwidth,height=60px]{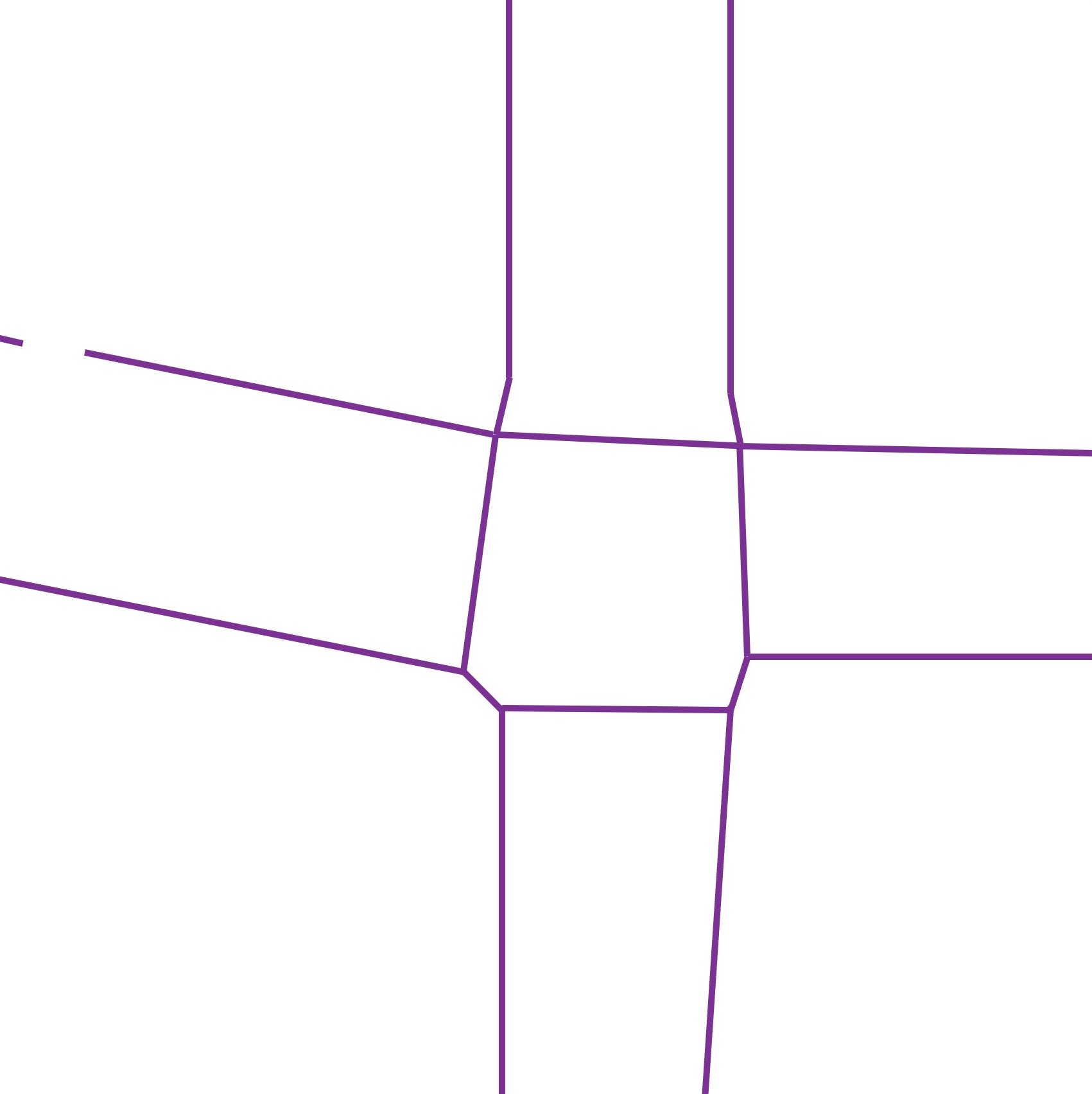} &
            \includegraphics[width=0.2\textwidth,height=60px]{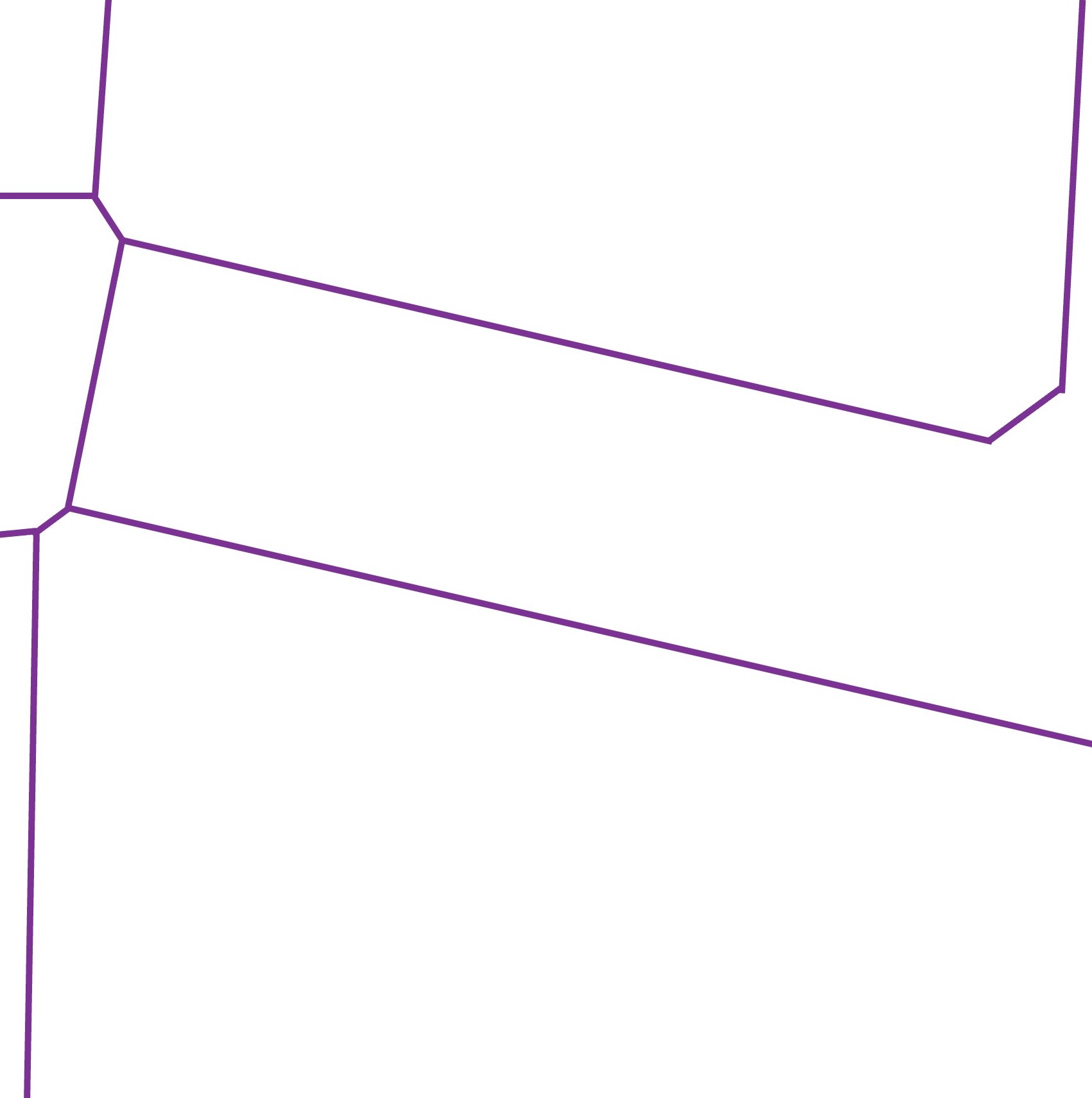} &
            \includegraphics[width=0.2\textwidth,height=60px]{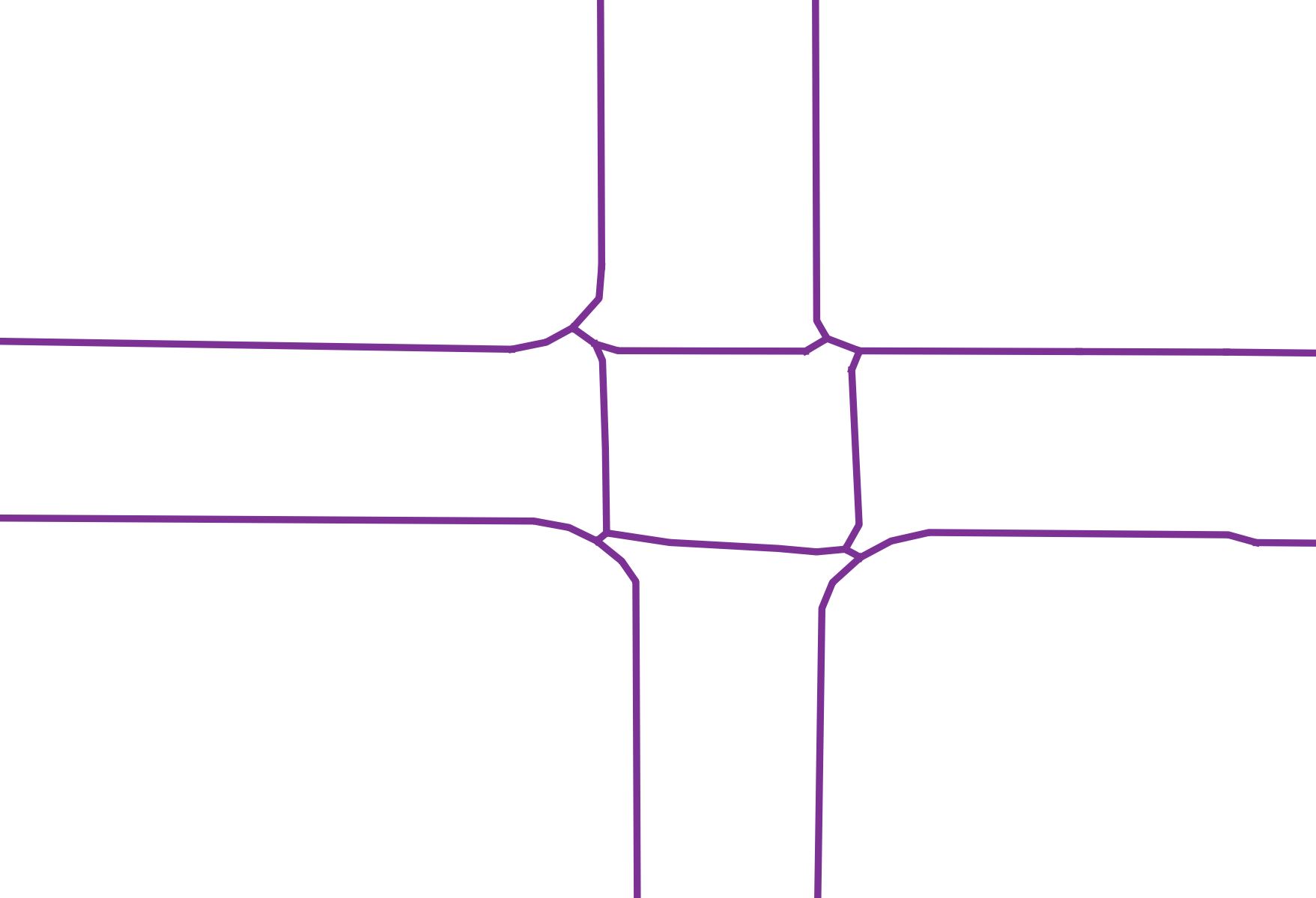} &
            \includegraphics[width=0.18\textwidth,height=60px]{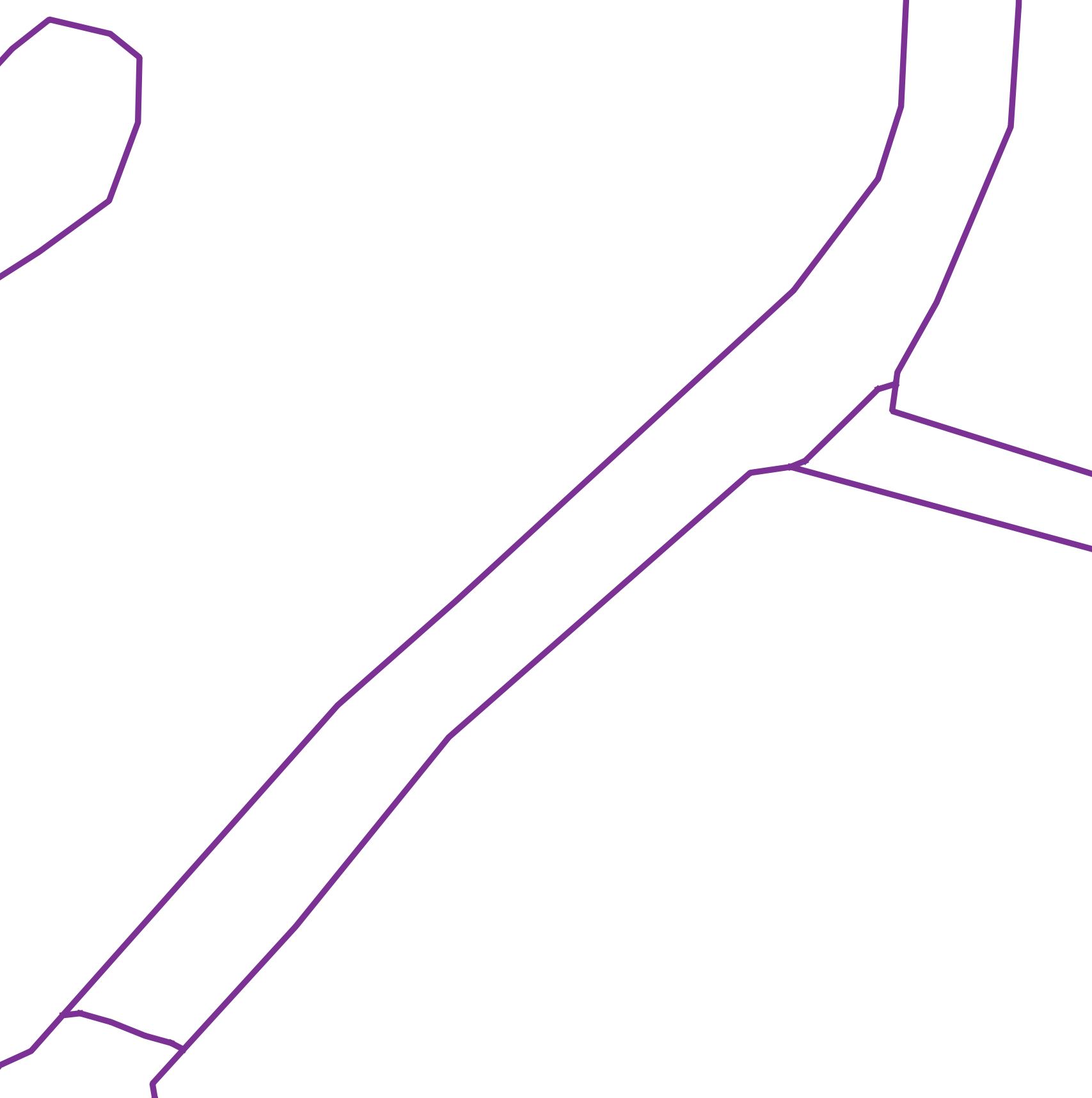} &
            \includegraphics[width=0.2\textwidth,height=60px]{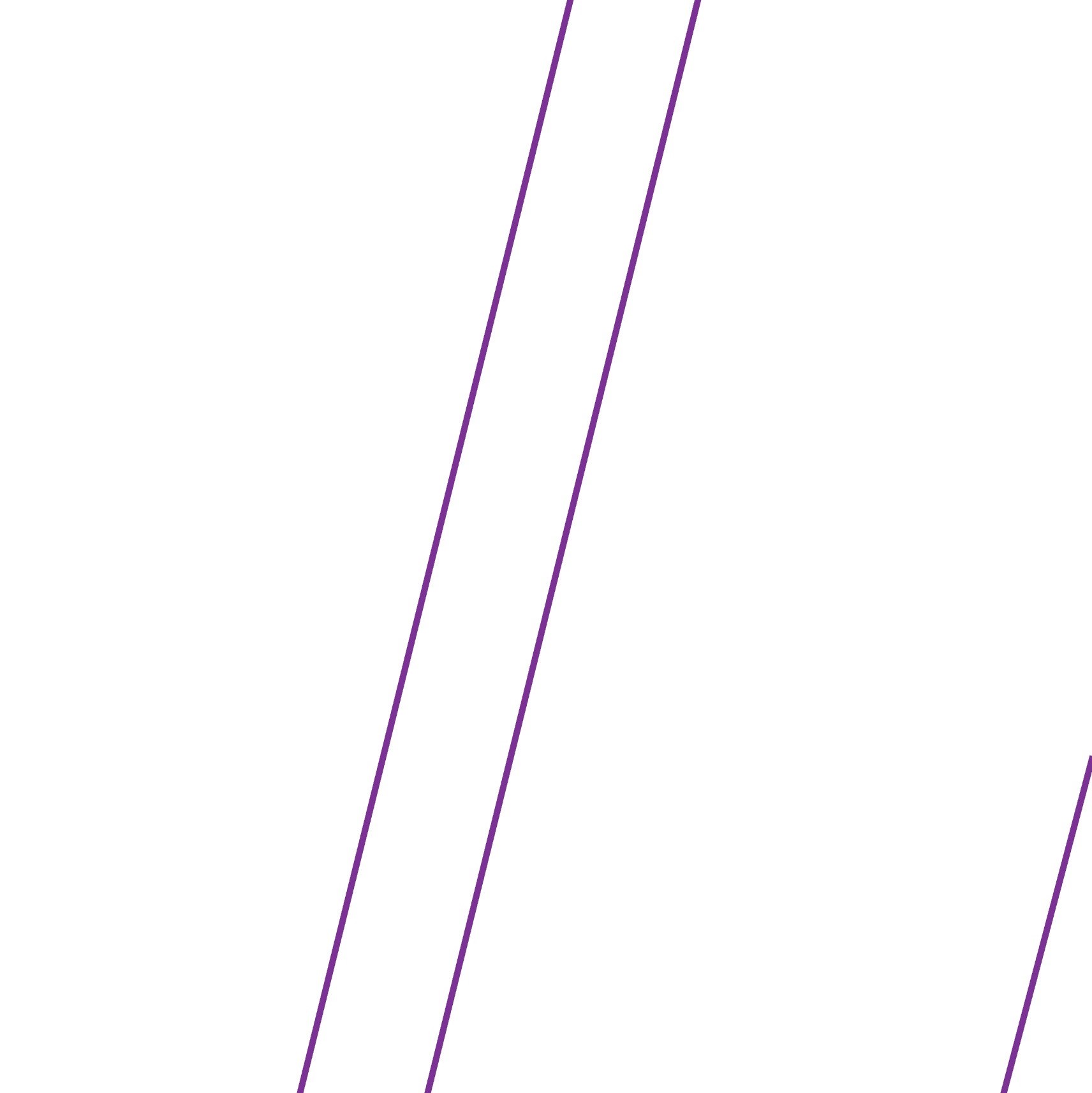}
            \\

        \end{tabular}
    }
    \caption{Samples from the PathwayBench dataset: imagery and annotations from various geographic areas offer great coverage and diversity.}
    \label{fig:data_sample}
\end{figure}

In addition to the graph annotation in GeoJSON, the PathwayBench dataset provides semantic segmentation annotations for three distinct classes needed to semantically segment the pedestrian environment, as many methods rely on segmentation as an intermediate step \cite{hosseini2023mapping, zhang2023ape}, and some methods produce only segmentations as output \cite{wu2019road}. Shown in Figure \ref{fig:data_sample}, these classes include (1)  \colorbox{corner}{\textcolor{classtext}{\textit{Corner bulb}}}  (2) \colorbox{sidewalk}{\textcolor{classtext}{\textit{Sidewalk}}}  (3) \colorbox{crossing}{\textcolor{classtext}{\textit{Crossing}}}. Corner bulbs are commonly used when describing a transportation network since they serve as a transition zone connecting a sidewalk to curb ramps, crossings, or another sidewalk. The nodes representing sidewalk endpoints, link endpoints, and curbs are usually located within the corner bulbs. Sidewalks and crossings are essential elements in an urban pedestrian path network graph, as the lines representing sidewalks and crossings are essentially the edges a pedestrian will traverse. The focus of our work is on mapping pedestrian paths, thus, all other annotated classes (including roads, buildings, and trees) collectively comprise the \textit{background} class in the experiments presented in this paper. However, these additional classes are also available in the PathwayBench dataset.


\subsection{Challenges}
\label{sec:data_challenge}
Imagery dataset bias can be introduced by non-representative geographic locations of images \cite{torralba2011unbiased, wilson2019predictive, jo2020lessons}. This bias generally results from a training set representing a limited geographic region or an environment that lacks diversity (e.g., trees, sidewalk materials, width, buildings). To expand PathwayBench's ability to generalize to other regions and environments, we designed a balanced and representative dataset that includes images from both North and South American urban contexts. However, only lower-resolution aerial satellite images were openly available for South American cities, which introduces an additional challenge for model learning. Figure \ref{fig:data_sample} displays samples from the PathwayBench dataset, with the first three columns representing North American cities and the last two columns representing lower-resolution images in South American cities. 

Occlusion represents an additional challenge that varies by region~\cite{ning2022sidewalk}. For example, in the third column shown in Figure \ref{fig:data_sample}, part of the pixels that are labeled as \textit{sidewalk} are occluded by buildings, the shade from buildings, and vegetation in the aerial images. In this case, learning from aerial images alone is challenging and street map imagery tiles can provide crucial auxiliary information. Section \ref{sec:exp} shows that using multiple sources of inputs improves the segmentation outcome.


\section{Routability and Traversability}
\label{sec:tiletraversability}


Routability could be measured by the precision and recall of all paths in the built environment, but the number of paths grows exponentially in the size of the graph, and not all paths are equally important\cite{liu2022generalized,bolten2021towards}.  Selecting specific paths between points of interest would be a subjective, city-specific decision.

A key observation is that high-quality global routes are constructed from high-quality local routes. To measure local routability, we impose polygons on the dataset and compare the local graph properties within each polygon to the ground truth, averaging the results (see teaser image). We consider four measures of polygon-based routability: 1) degree centrality, 2) betweenness centrality, 3) the number of connected components of the local polygon graph, and 4) traversability, a novel metric that measures the ability to travel through small intersection-scale regions and affords meaningful comparisons across arbitrary graphs by imposing no requirement that they share nodes or edges.

\paragraph{Problems with count-based metrics} Count-based metrics check for correspondence between edges in the ground truth and edges in the prediction. For example, \citet{hosseini2023mapping} reports the proportion of ground truth edges that are within 4 meters of at least one predicted edge.  This metric captures some notion of coverage but can generate high scores even when there is little connectivity in the graph (and therefore little utility for downstream applications), and can potentially generate low scores even when both graphs support very similar paths.


Ideally, we would measure precision and recall of all possible paths through the ground truth graph.  But measuring this notion of global routability is computationally infeasible given the size of a city-scale graph, and small (within-intersection) deviations in a global path are less noticeable to a traveler than large (different intersection) deviations; a traveler might have multiple ways to navigate an intersection and still consider themselves on the same global route.

\begin{wrapfigure}{r}{0.58\textwidth}
\vspace{-2em}
  \begin{center}
    \includegraphics[width=0.55\textwidth]{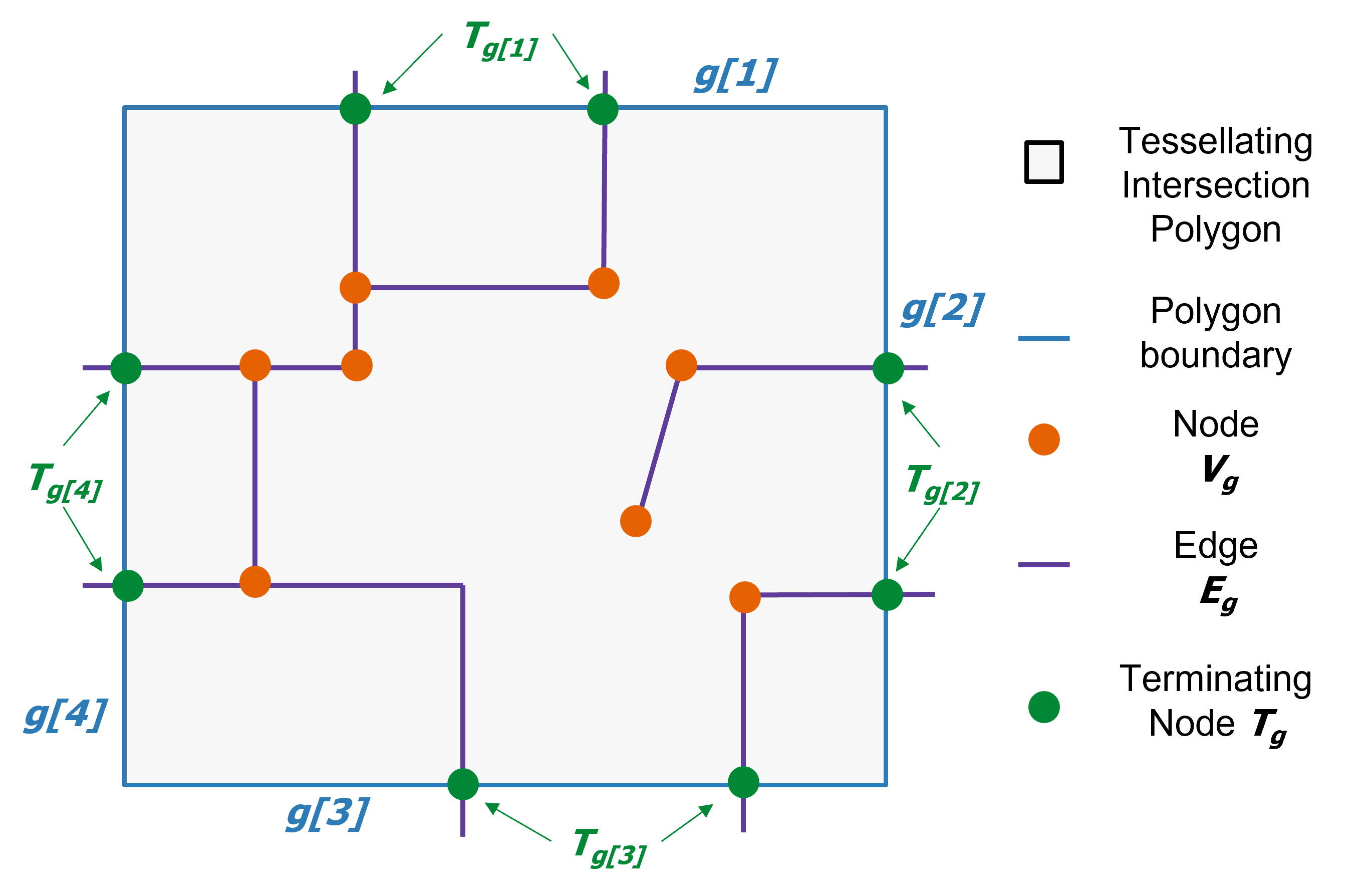}
  \end{center}
  \caption{Traversability measures the ability to navigate from one boundary of a polygon to another using the graph, favoring global routability over local variations. Boundary pair $g[4]$ and $g[3]$ is traversable; $g[1]$ and $g[2]$ are not.}
  \label{fig:traversability_annotated}
  \vspace{-0.4em}
\end{wrapfigure}

\paragraph{Intuition for traversability}
We address both problems by tesselating the space with polygons, about the size of one city block, but centered at intersections, and reasoning about boundary-to-boundary traversal within each polygon.  By using the same tesselation for both ground truth and prediction, we can directly compare results and derive a similarity metric. By computing boundary-to-boundary traversal only locally, we avoid enumerating all possible global paths, while still ensuring some correspondence between the paths afforded in ground truth and the paths afforded in the prediction. The intuition is that as long as a predicted path involves the same polygons as a ground truth path, we consider the ground truth path accurately represented, even if there are local, within-polygon variations.

Within a single polygon, the key idea is to assess a traveler's ability to move from one boundary of the polygon to another.  The reason this boundary-to-boundary traversal is important is that it can be defined on \textit{any} graph: If the ground truth graph affords traversal from the north boundary of (say) an intersection to the east boundary of the intersection, then a predicted graph should also afford such a traversal, and vice versa. Within a polygon, we can check each pair of boundaries ($boundary_1$, $boundary_2$) to see if any point on $boundary_1$ is reachable from any point on $boundary_2$.  If so, we say the boundary pair $(boundary_1, boundary_2)$ is traversable.  We then compare this set of traversable boundary pairs with the corresponding set of traversable boundary pairs in the ground truth using Jaccard similarity.  A high score indicates agreement between prediction and ground truth for that specific polygon: a traveler can enter and exit the polygon in the same ways on both graphs.

\paragraph{Definition: pedestrian graph}
Let \( G = (V, E, f) \) be a spatial \textit{pedestrian pathway graph} representing the entire city region, where $f : V \rightarrow \mathbb{R}^2$ assigns the spatial location of each node in $G$ for a 2D geometric space $\mathbb{R}^2$.  

\paragraph{Definition: terminating nodes, polygon graph} Let $g$ be a polygon in $\mathbb{R}^2$ where $g[i]$ is the $i$th boundary. Let $B$ be the set of edges $(x,y) \in E$ such that $x$ is inside $g$ and $y$ is outside $g$. For each boundary $g[i]$, let $T_{gi}$ be the set of points where some edge $b \in B$ geometrically intersects the boundary $g[i]$. We call these points of intersection \textit{terminating nodes}. Let $T_g$ be the set of all terminating nodes for any boundary in $g$. 

Figure \ref{fig:traversability_annotated} illustrates the situation.  There are four boundaries numbered $g[1],g[2],g[3],g[4]$, and four corresponding sets of terminating nodes, $T_{g[1]}, T_{g[2]}, T_{g[3]}, T_{g[4]}$.  Let $S_p$ be a \textit{polygon graph} $(V_g \cup T_g, E_g \cup B)$, where $V_g$ is the set of nodes from $G$ contained in the polygon, $E_g$ is the set of edges in $G$ contained in the polygon $g$ and $B$ is the set of "cut" edges $(x,t)$ where $t \in T_g$.

\paragraph{Definition: polygon traversability}
We then say that a pair of boundaries $(g[i], g[j])$ is \textit{traversable} in $G$, written $\textit{traversable}(g[i], g[j], G)$, if there exists a pair of nodes $(u,v)$ such that $u \in T_{gi}$, $v \in T_{gj}$, $u \neq v$, and $u$ is reachable from $v$ in the polygon graph $S_g$.  For example, in Figure \ref{fig:traversability_annotated}, the pair of boundaries $(g[1], g[4])$ is traversable, but the pair of boundaries $(g[1],g[2])$ is not. The other traversable boundary pairs are $(g[1], g[3])$, $(g[3], g[4])$, $(g[2], g[3])$, as well as $(g[1], g[1])$ and $(g[4], g[4])$, as we also consider the case where a boundary is traversable to itself.


The traversable boundary pairs of $g$ in $G$ is: 
\[
 \textit{TraversablePairs}(g, G) = \{ (g[i], g[j]) \; | \; i \in 1..|g|, j \leq i, \textit{traversable}(g[i], g[j], G) \}
\]



\paragraph{Traversability Metric}
Given a predicted pedestrian graph $P$ and a ground truth pedestrian graph $G$, we can compare the sets of traversable boundary pairs for a polygon from each graph using Jaccard similarity.  Specifically, given a polygon $g$:

\[
 \textit{TraversabilitySimilarity}(g, G, P) = \frac{\lvert\textit{TraversablePairs}(g, G) \cap \textit{TraversablePairs}(g, P)\rvert}{\lvert\textit{TraversablePairs}(g, G) \cup \textit{TraversablePairs}(g, P)\rvert}
\]

Given a predicted graph $P$, a ground truth graph $G$, and a polygon $g$, if a boundary pair $(g[i], g[j])$ is traversable in $G$ but not in $P$, then all global paths in $G$ that cross $g[i]$ and $g[j]$ will not be present in $P$.  These path deviations are significant in the sense that they do not involve the same sequence of polygons.  In practice, this means that routes in the predicted graph will typically involve different intersections, which can be interpreted as a failure of the model to represent global paths. 


\paragraph{Implementation}
To represent traversability, we first partition the entire test area into Tessellating Intersection Polygons (TIP). Each TIP is created by assigning a point location to a road intersection, then computing the associated Voronoi polygons to tessellate the entire test area.  While any tesselation can be used with the metric, centering each TIP on an intersection avoids trivial TIPs that have little effect on global pedestrian routes.

The TIPs can also be used with other metrics. For example, a predicted graph should tend to agree with a ground truth graph on the number of connected components, the average degree, or the average betweenness centrality (BC).  However, even the popular Brandes approximation algorithm~\cite{brandes2001faster} for BC is quadratic ($O(nm + n2logn)$ for $n$ nodes and $m$ edges). By computing betweenness centrality locally within a TIP, we capture a notion of average local connectedness while remaining computationally efficient.

\section{Experiments}
\label{sec:exp}

\subsection{Segmentation Assessment with Multiple Inputs}
\label{sec:exp_train}
\begin{figure}[ht!]
    \centering
    \resizebox{0.9\columnwidth}{!}{
    \begin{tabular}{c|c|c|c|c|c|c}
     \textbf{Aerial Image} & \textbf{Rasterized street map} & \textbf{Ground Truth}  & \textbf{Satellite only} & \textbf{Street only} & \textbf{Satellite + Street} & \textbf{Prediction Graph}\\
        \includegraphics[width=0.33\columnwidth]{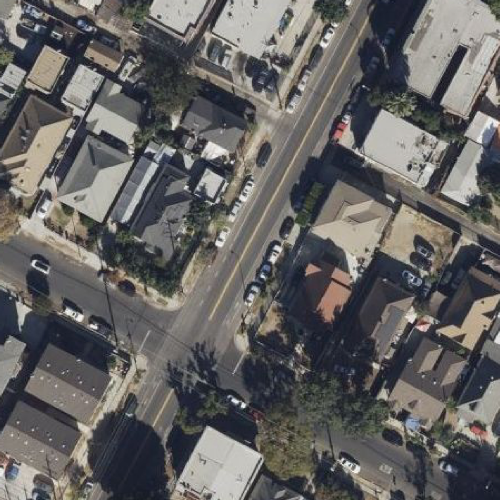}   &
        \includegraphics[width=0.33\columnwidth]{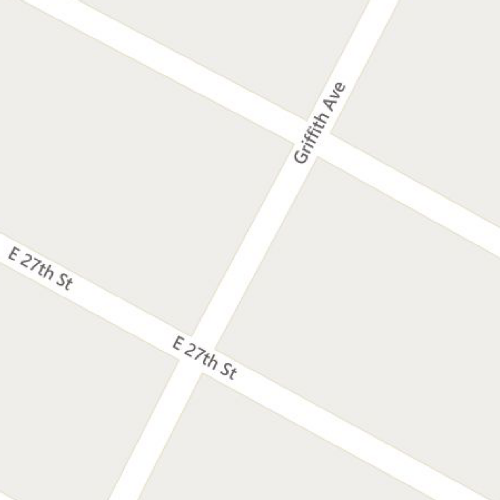} &
         \includegraphics[width=0.33\columnwidth]{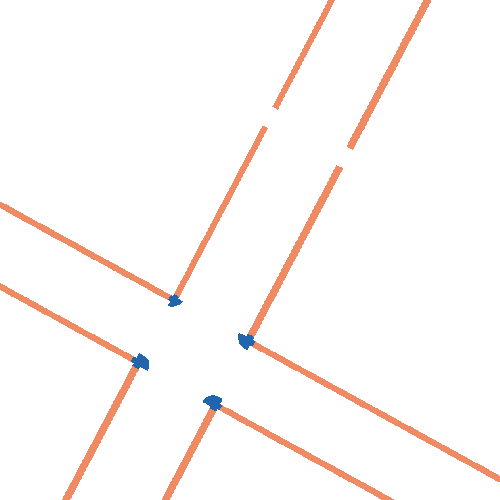} &
        \includegraphics[width=0.33\columnwidth]{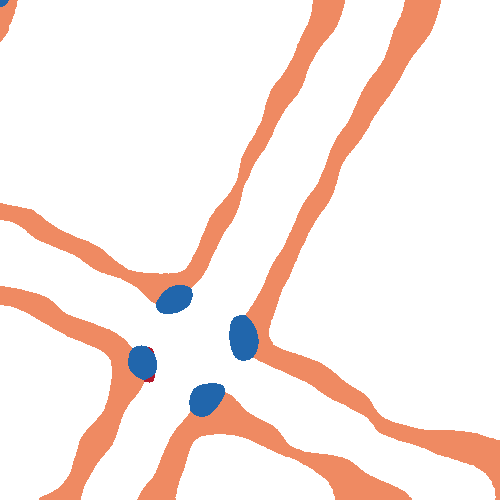} &
        \includegraphics[width=0.33\columnwidth]{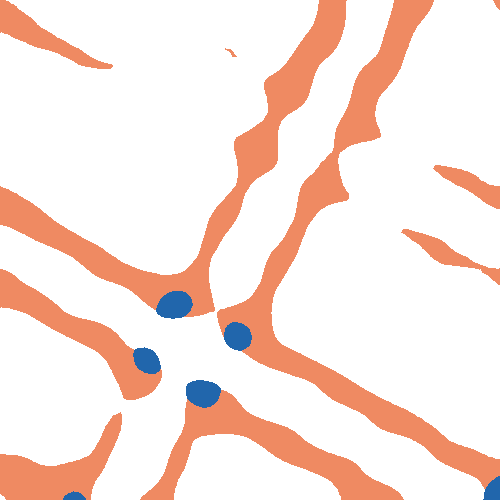} &
        \includegraphics[width=0.33\columnwidth]{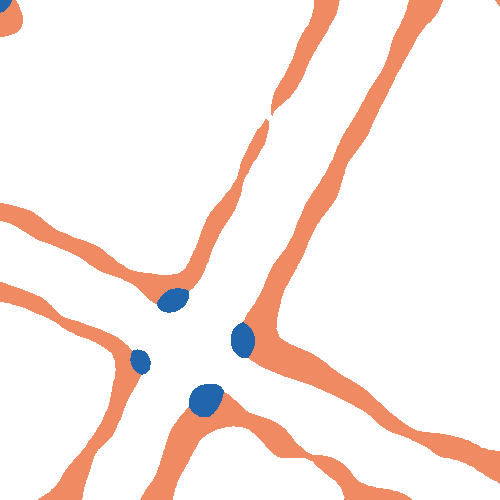} &
        \includegraphics[width=0.33\columnwidth]{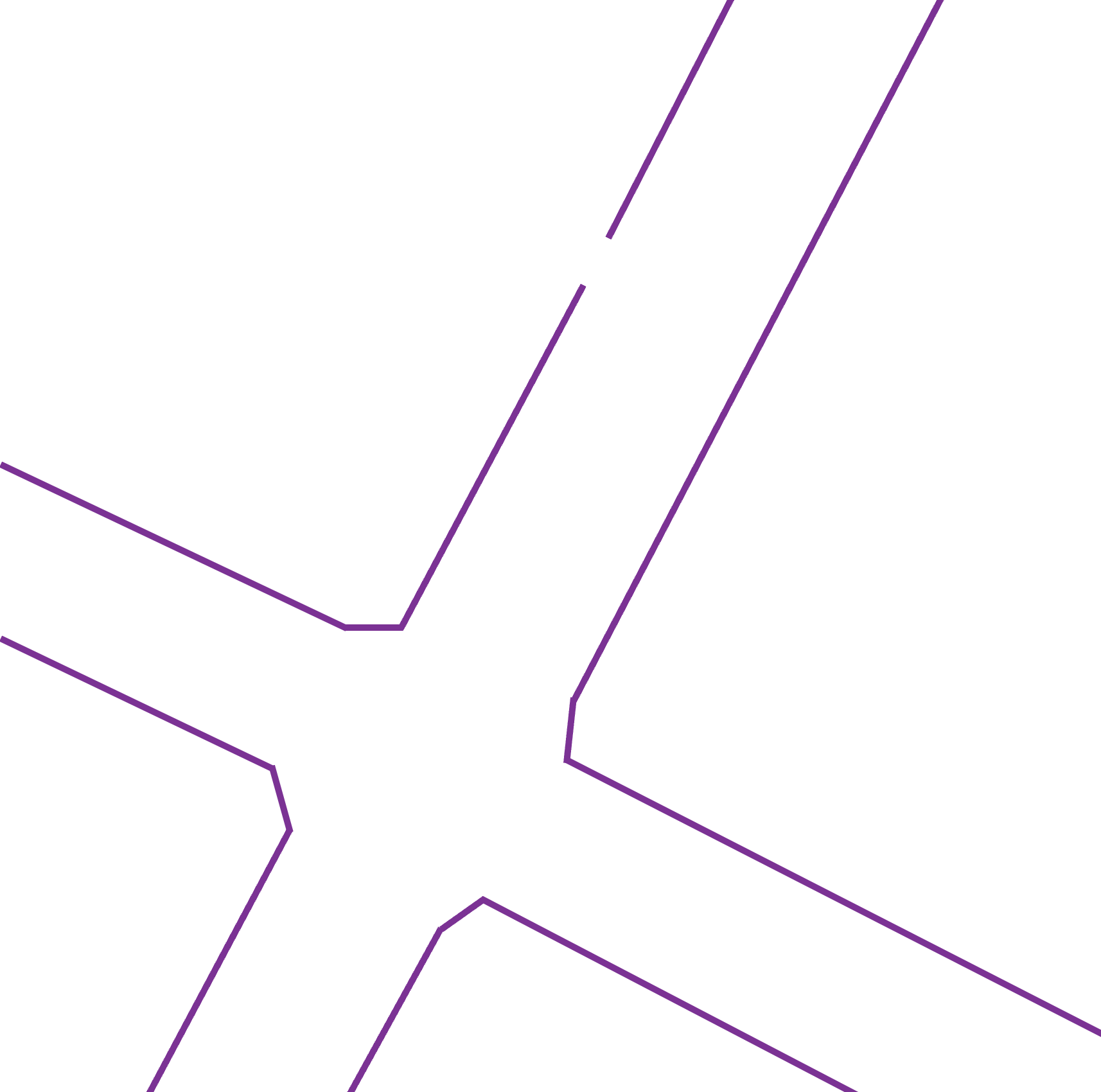} 
        \\
        \includegraphics[width=0.33\columnwidth]{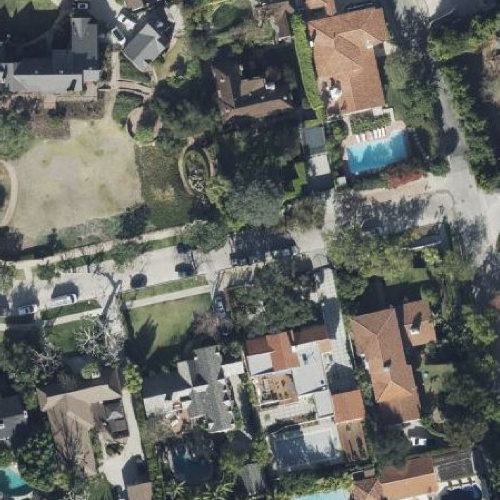}   &
        \includegraphics[width=0.33\columnwidth]{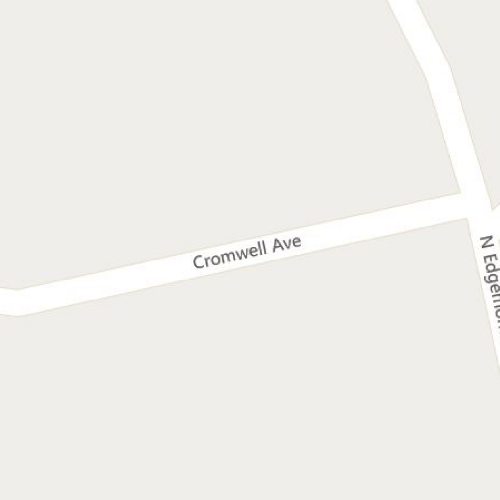} &
         \includegraphics[width=0.33\columnwidth]{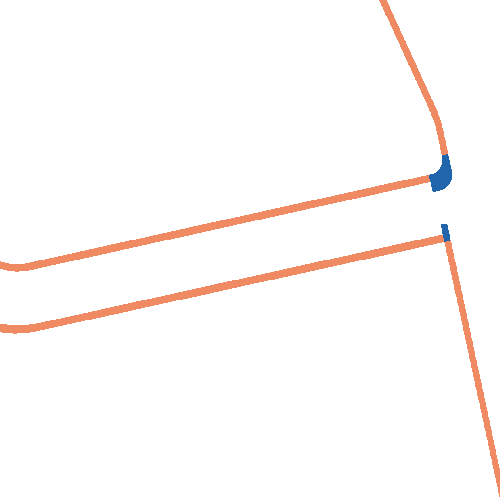} &
        \includegraphics[width=0.33\columnwidth]{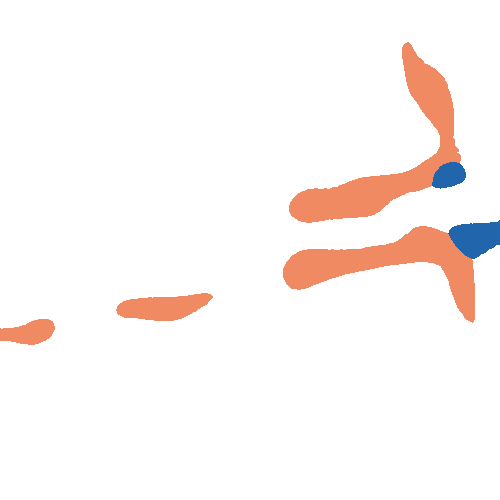} &
        \includegraphics[width=0.33\columnwidth]{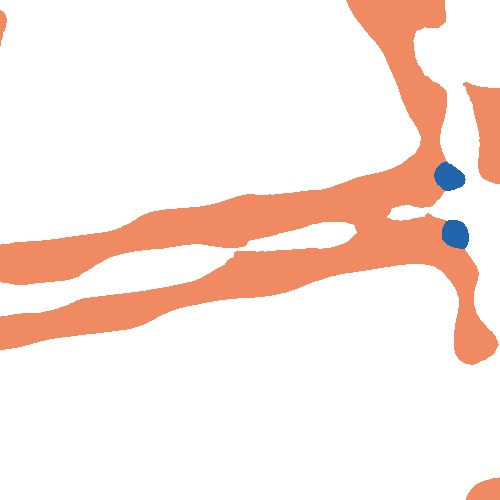} &
        \includegraphics[width=0.33\columnwidth]{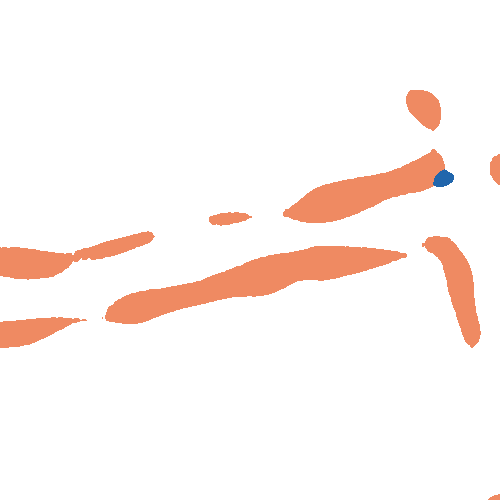} &
        \includegraphics[width=0.33\columnwidth]{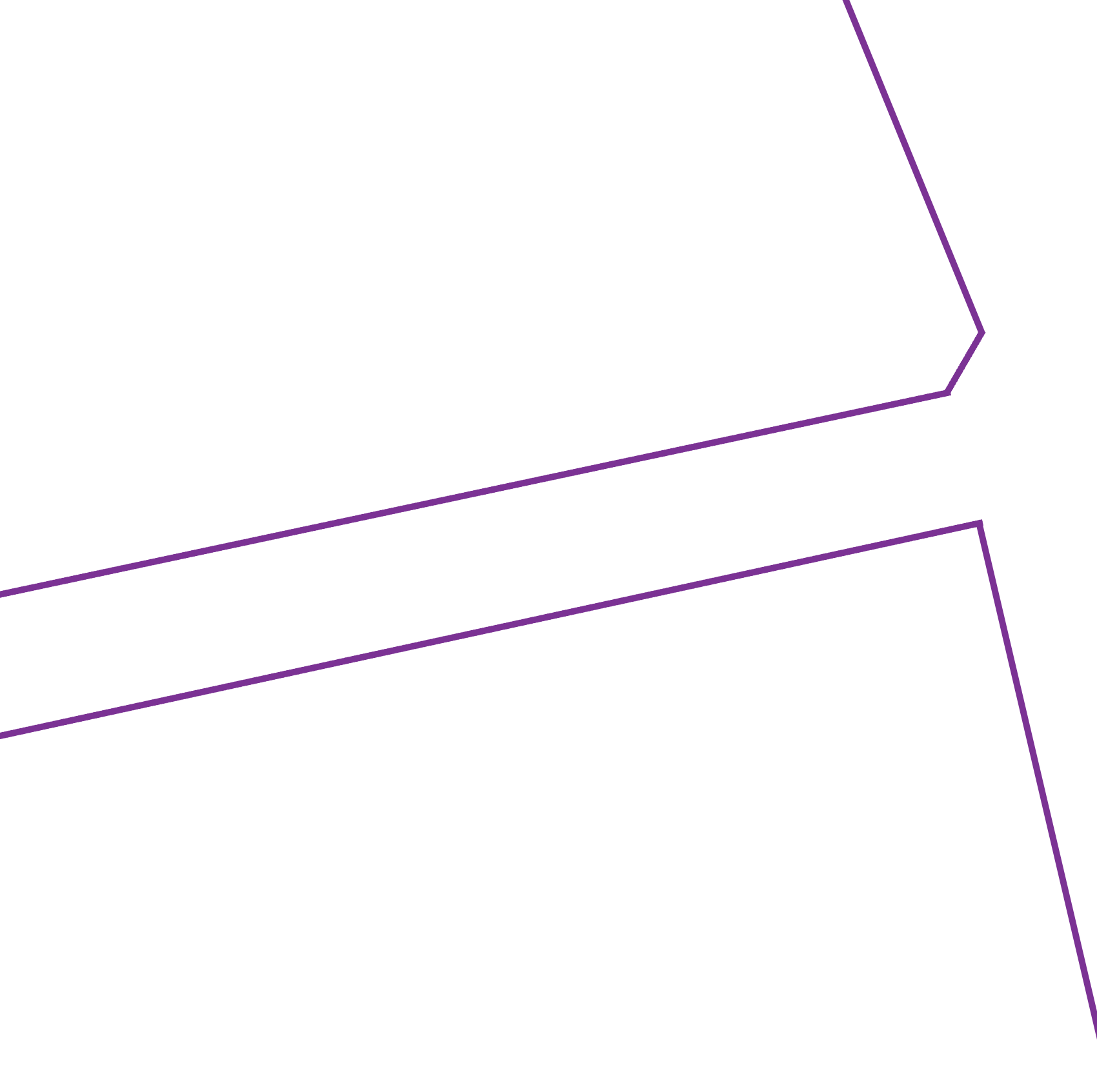} 
        \\
        \includegraphics[width=0.33\columnwidth]{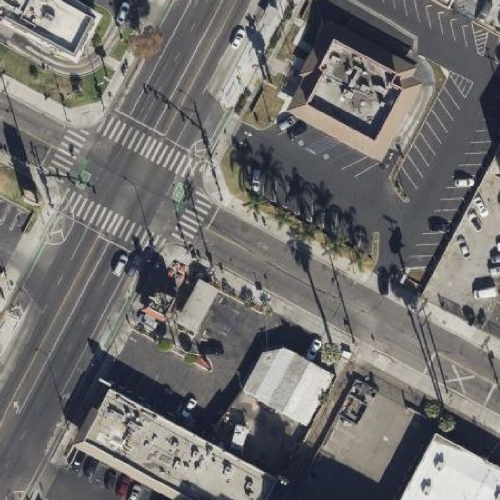}   &
        \includegraphics[width=0.33\columnwidth]{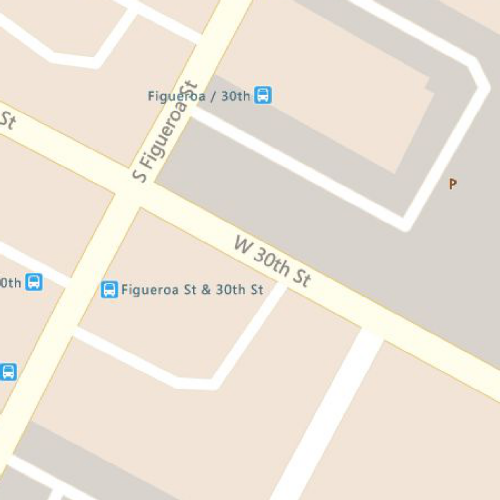} &
         \includegraphics[width=0.33\columnwidth]{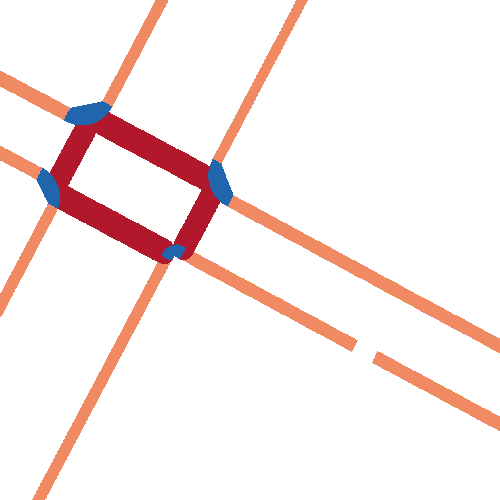} &
        \includegraphics[width=0.33\columnwidth]{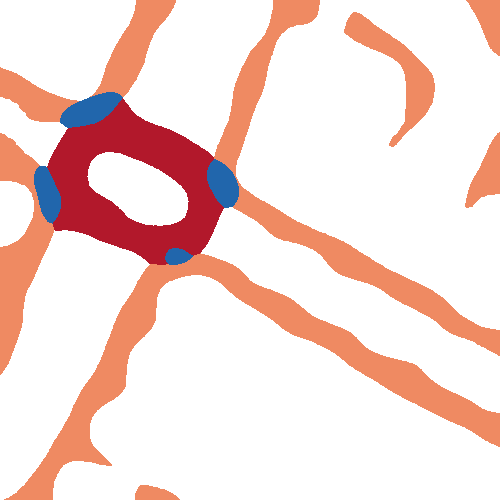} &
        \includegraphics[width=0.33\columnwidth]{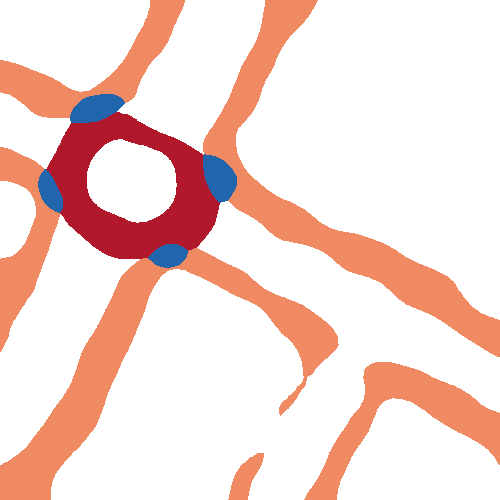} &
        \includegraphics[width=0.33\columnwidth]{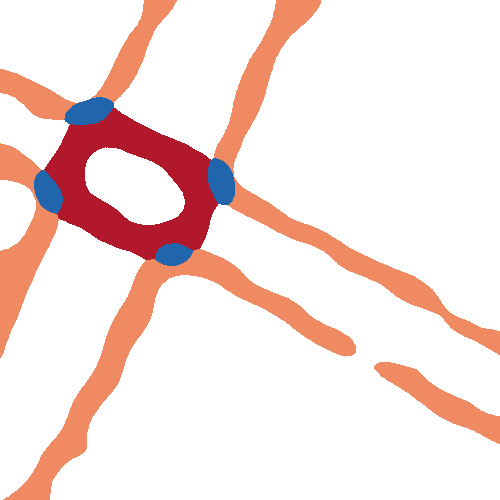} &
        \includegraphics[width=0.33\columnwidth]{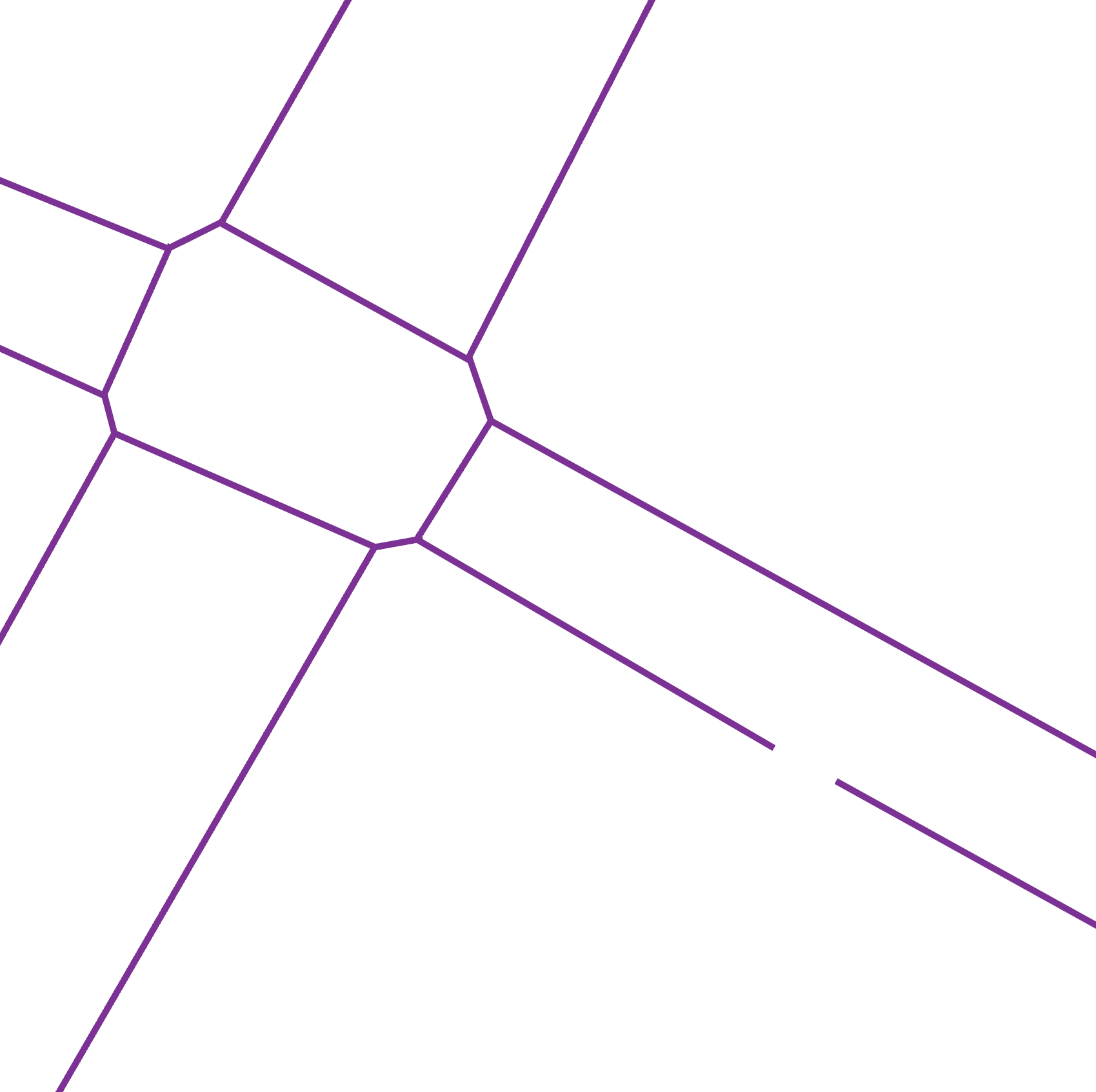} 
        \\
 \end{tabular}
 }
    \caption{Qualitative assessment on segmentation. The model that uses both aerial satellite images and street map images generates better predictions than the models that use only one input.}
    \label{fig:qual_samples}
\end{figure}

As many methods rely on segmentation as an intermediate step \cite{hosseini2023mapping, zhang2023ape} we provide annotated segmentations as part of the benchmark. We consider three neural architectures for segmentation:
an aerial satellite image branch only, a street map image tile branch only, and both aerial and street map image tile branches. All models trained used the same dataset split and data augmentation techniques. Performance comparisons are shown in Figure \ref{fig:qual_samples}. We train and validate with an $80\%/20\%$ split of the PathwayBench dataset. We augment the data by rotating, cropping, and resizing to improve generalization. 

Figure \ref{fig:qual_samples} visualizes the segmentation results on the validation set. The segmentation for \textit{sidewalk}, \textit{crossing}, and \textit{corner bulb} align well with the ground-truth segmentation. These qualitative examples also show the difficulty in predicting pedestrian path network classes with single-source input and the improvement gained from adding other input sources.
In the example shown in the first row, the prediction made with the model trained with only rasterized street map generated false-positive \textit{sidewalk} predictions. Adding aerial images during model training helped remove these spurious predictions. Predictions made with only aerial images suffered from occlusion vegetation, generating sparse, disconnected graphs. Including street image tiles during the training and inference helped recover many occluded sidewalks. Quantitative segmentation evaluation is in the supplemental material, and a quantitative analysis of the performance at the graph connectivity and routability level for different methods when testing in different cities is provided in Section \ref{sec:eval_graph}.

\subsection{Pathway Network Graph Routability Analysis}
\label{sec:eval_graph}
Pixel-wise measures do not reflect the routability of the predicted graph. mIoU (or other pixel-wise measures) cannot measure how close a predicted graph is to the ground truth graph. To measure routability, we use graph-level metrics we proposed in Section \ref{sec:tiletraversability}. The methods we included for the evaluation are: (1) \textbf{Tile2Net} \cite{hosseini2023mapping}: This method uses aerial satellite images to segment sidewalks, crosswalks, and footpaths in cities. It then simplifies the segmented polygons, and extracts the centerlines of the polygons to represent pathway graphs. (2) \textbf{Pedestrianfer} \cite{zhang2023ape}: This method uses the street network graph to heuristically infer nearby pathway edges, under the assumption that most streets have associated sidewalks. (3) \textbf{Prophet} \cite{zhang2023ape}: This method uses Pedestrianfer to generate a hypothesis,  then uses the rasterized street map and aerial imagery to generate the segmentation results (with different backbones VGG-16~\cite{simonyan2014very}, DeepLabv3~\cite{chen2017rethinking}, and ViT~\cite{dosoViTskiy2020} ) to infer the sidewalks, crossings, and corner bulbs, and finally uses the predicted segmentation mask to refine, correct, and optimize the hypothesized graph. 

\begin{minipage}[htp!]{0.99\textwidth}
    \centering
    \captionsetup{width=\textwidth}
    \captionof{table}{Pathway Network Graph Routability Evaluation}
    \label{tab:graph_eval}
    \resizebox{0.99\textwidth}{!}{%
    \begin{tabular}{l|l|ccc|cc|cc}
    \toprule
    \multirow{2}{*}{\textbf{Method}} & \multirow{2}{*}{\textbf{Area}} & \multicolumn{3}{c|}{\textbf{Global}} & \multicolumn{2}{c|}{\textbf{Local}} & \multicolumn{2}{c}{\textbf{Local (relative to Ground Truth)}}\\
    & & $\textbf{\# nodes}$ & $\textbf{\# edges}$ & $\textbf{avg degree}$ & $\textbf{avg CC}$ & $\textbf{avg BC}$ & $\textbf{edge-retrieval F1}$ & $\textbf{\textit{TraversabilitySimilarity}}$\\ 
    \midrule
     & Washington, D.C. & 19245 & 21462 & 2.23 & 1.62 & 0.14 & 1.0 & 1.0\\ 
    Ground Truth         & Portland, OR  & 937 & 1134 & 2.42 & 1.45 & 0.13 & 1.0  & 1.0\\
             & Seattle, WA & 9386 & 10402 & 2.22 & 1.65 & 0.13 &  1.0 & 1.0\\ 
    \midrule
     & Washington, D.C. & 27651 & 24948 & 1.80 & 6.44 & 0.02 & 0.84 & 0.37\\ 
     Pedestrianfer             & Portland, OR & 1615  & 1376 & 1.70 & 6.46 & 0.02 & 0.89 & 0.30\\ 
                  & Seattle, WA & 10002  & 8871 & 1.77 & 5.43 & 0.03 &  \textbf{0.93} & 0.38\\
    \midrule
     & Washington, D.C. & 48153 & 46369 & 1.93 & 5.77 & 0.03 & 0.84 & 0.35\\ 
    Tile2net        & Portland, OR & 937  & 844 & 1.80 & 2.77 & 0.04 & 0.76 & 0.04\\ 
            & Seattle, WA & 12617 & 11732 & 1.86 & 5.01 & 0.03 & 0.90  & 0.17
            \\
    \midrule
     & Washington, D.C. & 8908 & 10397 & 2.33 & 1.15 & 0.13 & 0.84 & 0.38\\ 
    Prophet(VGG-16) 
        & Portland, OR & 945 & 1056 & 2.23 & \textbf{1.39} & 0.13 & 0.90 & 0.47\\ 
        & Seattle, WA & 5061 & 5497 & 2.17 & 1.37 & 0.13 & 0.89 & 0.41\\
    \midrule
          & Washington, D.C. & 9967 & 11503 & 2.22 & 1.12 & 0.13 & 0.87 & 0.39 \\
    Prophet (DeepLabv3)   
         & Portland, OR & 1138 & 1267 & 2.25 & 1.71 & 0.14 & 0.87 & 0.48\\ 
         & Seattle, WA & 5472 & 6046 & 2.21 & 1.35 & 0.14 & 0.89 & 0.42\\
    \midrule

         & Washington, D.C. & 10347 & 11608 & \textbf{2.22} & \textbf{1.20} & \textbf{0.13} & \textbf{0.88} & \textbf{0.39}\\
    Prophet (ViT)
         & Portland, OR & 1121 & 1205 & \textbf{2.26} & 1.68 & \textbf{0.13} & \textbf{0.91}  & \textbf{0.48}\\ 
         & Seattle, WA & 5516 & 6247 & \textbf{2.22} & \textbf{1.39} & \textbf{0.13} & 0.90 & \textbf{0.43}\\
    \bottomrule
    \end{tabular}
    }
\end{minipage}

The metrics considered are (1) local, polygon-level routability metrics described in Section \ref{sec:tiletraversability} (average number of connected components (avg CC), the average betweenness centrality (avg BC), and \textit{TraversabilitySimilarity}),  (2) count-based metrics (node count, edge count, and avg degree for the entire test area), and (3) the F1 scored based on the network segment edge-retrieval method described in previous work \cite{hosseini2023mapping}. The evaluation results are summarized in Table \ref{tab:graph_eval}. 

\textit{TraversabilitySimilarity} compares whether local intersections afford the same ingress and egress points in both predicted and ground truth graphs.  We include avg CC and avg BC as additional measures of local connectivity, where the distance from ground truth values exposes deviations from the ground truth graph structure. Tile2Net, despite robust results in terms of the number of edges and nodes, and relatively high edge-retrieval F1 scores across all test areas, shows considerable difference in avg CC and avg BC values across all cities when compared to the ground truth. Tile2Net tends to overpredict disconnected edges, generating more connected components and lower average betweenness centrality. Moreover, \textit{TraversabilitySimilarity} scores of Tile2Net are remarkably low  (0.04 in Portland and 0.17 in Seattle), suggesting that global paths using the Tile2Net graph would involve significant deviations from ground truth, i.e., they would tend to route pedestrians through completely different intersections. Conversely, the Prophet models, particularly those utilizing the ViT architecture, show better performance in graph routability. They achieve the highest \textit{TraversabilitySimilarity} values among the evaluated models (0.48 and 0.43 for Portland and Seattle respectively), despite the F1 scores not being significantly higher than those of other models. This finding highlights a critical insight: while the F1 score captures similarity between predicted and ground truth graphs in terms of edge accuracy, it does not reflect how travelers navigate the environment. The F1 score is therefore less useful for downstream applications that rely on routing. \textit{TraversabilitySimilarity} better aligns with the practical requirements of real-world routing and navigation tasks, unlike count-based metrics used in prior work.

\section{Conclusion}
\label{sec:dis}
The PathwayBench benchmark provides a diverse, challenging,  practical, and comprehensive evaluation mechanism for extracting pedestrian pathways from aerial imagery, improving on the inconsistency in both datasets and metrics used in the literature. PathwayBench includes five co-registered features useful for inferring high-accuracy pathway graphs, along with a suite of metrics tailored to assess the structure of the graph to ensure the routability properties needed by downstream applications. Our experiments demonstrate that PathwayBench effectively highlights the strengths and limitations of existing methods, which were obscured by conventional pixel-wise evaluations and simple edge-counting approaches. By focusing evaluation on the experience of travelers, we believe this work will catalyze advancements in the extraction of large-scale pedestrian pathway networks.

\begin{ack}
This work was funded in part by the Taskar Center for Accessible Technology, USDOT ITS4US NOFO No: 693JJ322NF00001, and Microsoft's AI4Accessibility award. Thanks to the G3ict organization for its support of the Quito, Sao Paulo, Santiago, and Gran Valparaiso, on-the-ground mapping efforts.
\end{ack}

\bibliographystyle{ACM-Reference-Format}
\bibliography{sample-base}

\end{document}